\documentstyle[aclap]{article}

\title{{\small Appears in the Proceedings of the 5th Conference on
Applied \\ Natural Language Processing (ANLP-97), April 1997, Washington, 
DC\\}
Sequential Model Selection for Word Sense Disambiguation
\thanks{This research was supported by the Office of Naval
Research under grant number N00014-95-1-0776.}}

\author{Ted Pedersen\dag \and Rebecca Bruce\dag \and Janyce Wiebe\ddag \\
\dag Department of Computer Science and Engineering\\
Southern Methodist University, Dallas, TX 75275 \\
\ddag Department of Computer Science\\
New Mexico State University, Las Cruces, NM 88003 \\
{\tt pedersen@seas.smu.edu, rbruce@seas.smu.edu, wiebe@cs.nmsu.edu}}

\begin{document}
\maketitle
       
\begin{abstract}
Statistical models of word--sense disambiguation are often based on a
small number of contextual features or on a model that is assumed to
characterize the interactions among a set of features. Model
selection is presented as an alternative to these approaches, where a
sequential search of possible models is conducted in order to find the 
model that best characterizes the interactions among features. This
paper expands existing model selection methodology and presents the
first comparative study of model selection search strategies and
evaluation criteria when applied to the problem of building
probabilistic classifiers for word--sense disambiguation. 
\end{abstract}

\section{Introduction}

In this paper word--sense disambiguation is cast as a problem in
supervised learning, where a classifier is induced from a corpus of
sense--tagged text. Suppose there is a training sample where each
sense--tagged sentence is represented by the feature variables $(F_1,  \ldots,
F_{n-1}, S)$. Selected contextual properties of the sentence are
represented by $(F_1,\ldots, F_{n-1})$ and the sense of the ambiguous
word is represented by $S$. Our task is to induce a classifier that will
predict the value of $S$ given an untagged sentence represented by the
contextual feature variables. 

We adopt a statistical approach whereby a probabilistic model is
selected that describes the interactions among the feature
variables. Such a model can form the basis of a probabilistic
classifier since it specifies the probability of observing any and all
combinations of the values of the feature variables. 

Suppose our training sample has $N$ sense--tagged sentences.  
There are $q$ possible combinations of values for the
$n$ feature variables, where each such combination is represented by a
feature vector. Let $f_i$ and $\theta_i$ be the frequency and probability  of
observing the $i^{th}$ feature vector, respectively. Then
$(f_1,\ldots, f_q)$ has a multinomial distribution with parameters
$(N,\theta_1, \ldots, \theta_q)$. 
The $\theta$ parameters, i.e., the joint parameters, define the 
joint probability distribution of the feature variables.  These
are the parameters of the fully saturated model, the model in
which the value of each variable directly affects the values of all
the other variables. These parameters can be estimated as maximum 
likelihood estimates (MLEs), such that the estimate of $\theta_i$,
$\widehat{\theta_i}$, is $\frac{f_i}{N}$. 

For these estimates to be reliable, each of the $q$ possible combinations 
of feature values must occur in the training sample.  This is unlikely
for NLP data samples, which are often sparse and highly skewed 
(c.f., e.g. \cite{PedersenKB96} and \cite{Zipf35}).   

However, if the data sample can be adequately characterized by a less
complex model, i.e., a model in which there are fewer interactions
between variables, then more reliable parameter estimates can be
obtained:  In the case of decomposable models (Darroch et al., 1980; see 
below), the parameters of a less complex model are parameters of marginal
distributions, so the MLEs involve frequencies of combinations of values
of only subsets of the variables in the model. How well a model 
characterizes the training sample is determined by measuring the {\it fit} 
of the model to the sample, i.e., how well the distribution defined by the 
model matches the distribution observed in the training sample.  

A good strategy for developing probabilistic classifiers is to
perform an explicit model search to select
the model to use in classification.  This paper presents
the results of a comparative study of search strategies
and evaluation criteria for measuring model fit.
We restrict the selection process to the class of decomposable models
\cite{DarrochLS80}, since restricting model search to this class has many
computational advantages. 

We begin with a short description of decomposable models (in section 2). 
Search strategies (in section 3) and model evaluation (in section 4)
are described next, followed by the results of an extensive
disambiguation experiment involving 12 ambiguous words (in sections
5 and 6). We discuss related work (in section 7) and close with 
recommendations for search strategy and evaluation criterion when
selecting models for word--sense disambiguation.  

\section{Decomposable Models}

Decomposable models are a subset of the class of graphical models
\cite{Whittaker90} which are in turn a subset of the class of
log-linear models \cite{BishopFH75}.  Familiar examples of
decomposable models are Naive Bayes and n-gram models. 
They are
characterized by the following properties \cite{BruceW94B}:       

\begin{enumerate}
\item In a graphical model, variables are either interdependent or 
conditionally independent of one another.\footnote{$F_2$ and $F_5$ are 
conditionally independent given $S$ if $p(F_2|F_5,S)$ = $p(F_2|S)$.} All 
graphical models have a graphical representation such that each variable 
in the model is mapped to a node in the graph, and there is an undirected 
edge between each pair of nodes corresponding to interdependent variables. 
The sets of completely connected nodes (i.e., cliques) correspond to sets 
of interdependent variables.  Any two nodes that are not directly 
connected by an edge are conditionally independent given the values of the 
nodes on the path that connects them.

\item Decomposable models are those graphical models that express the
joint distribution as the product of the marginal distributions of the 
variables
in the maximal cliques of the graphical representation, scaled by the
marginal distributions of variables common to two or more of these maximal
sets.  Because their joint distributions have such closed-form
expressions, the parameters can be estimated directly from the training
data without the need for an iterative fitting procedure (as is required,
for example, to estimate the parameters of maximum entropy models;
\cite{BergerDD96}). 

\item Although there are far fewer decomposable models than log-linear
models for a given set of feature variables, it has been shown that
they have substantially the same expressive power \cite{Whittaker90}. 
\end{enumerate}

The joint parameter estimate 
$\widehat\theta_{f_1,f_2,f_3,s_i}^{F_1,F_2,F_3,S}$
is the probability that the feature vector
($f_1,f_2,f_3,s_i$) will be observed in a training sample where each
observation is represented by the feature variables $(F_1,F_2,F_3,S)$. 
Suppose that the graphical representation of a decomposable model is
defined by the two cliques (i.e., marginals) $(F_1,S)$ and
$(F_2,F_3,S)$. The frequencies of these marginals,
$f(F_1=f_1,S=s_i)$ and  $f(F_2=f_2, F_3=f_3,S=s_i)$, are sufficient
statistics in that they provide enough information to calculate
maximum likelihood estimates of the model parameters. MLEs of
the model parameters are simply the marginal frequencies normalized by
the sample size $N$. The joint parameter estimate is formulated as a
normalized product:

\begin{equation}
\widehat\theta_{f_1,f_2,f_3,s_i}^{F_1,F_2,F_3,S} = 
\frac{\frac{f(F_1=f_1,S=s_i)}{N} \times
\frac{f(F_2=f_2,F_3=f_3,S=s_i)}{N}}
{\frac{f(S=s_i)}{N}}
\end{equation}

Rather than having to observe the complete feature vector
($f_1,f_2,f_3,s_i$) in the training sample to estimate the
joint parameter, it is only necessary to observe the marginals
$(f_1,s_i)$ and $(f_2,f_3,s_i)$. 
	
\section{Model Search Strategies}

The search strategies presented in this paper are backward sequential
search (BSS) and forward sequential search (FSS).  Sequential searches
evaluate models of increasing (FSS) or decreasing (BSS) levels of
complexity, where complexity  is defined by the number of interactions
among the feature variables (i.e., the number of edges in the
graphical representation of the model).   

A backward sequential search (BSS) begins by designating the saturated 
model as the current model. A saturated model has complexity level  
$i = \frac{n(n-1)}{2}$,  where $n$ is the number of feature variables. 
At each stage in BSS we generate the set of decomposable models of
complexity level $i-1$ that can be created by removing an edge from
the current model of complexity level $i$. Each member of this set is
a hypothesized model and is judged by the evaluation criterion to
determine which model results in the least degradation in fit from the
current model---that model becomes the current model and the search
continues.  The search stops when either (1) every hypothesized model
results in an unacceptably high degradation in fit or (2) the current
model has a complexity level of zero.

A forward sequential search (FSS) begins by designating the model of
independence as the current model. The model of independence has
complexity level $i=0$ since there are no interactions among the
feature variables.  At each stage in FSS we generate the set of
decomposable models of complexity level $i+1$ that can be created by
adding an edge to the current model of complexity level $i$. Each
member of this set is a hypothesized model and is judged by the
evaluation criterion to determine which model results in the greatest
improvement in fit from the current model---that model becomes the
current model and the search continues.  The search stops when either
(1) every hypothesized model results in an unacceptably small increase
in fit or (2) the current model is saturated.

For sparse samples FSS is a natural choice since early in the search the
models are of low complexity. The number of model parameters is small and 
they have more reliable estimated values. On the other hand, BSS begins 
with a saturated model whose parameter estimates are known to be 
unreliable. 

During both BSS and FSS, model selection
also performs feature selection. If a model is selected where
there is no edge connecting a feature variable to the classification
variable then that feature is not relevant to the  classification being
performed. 

\section{Model Evaluation Criteria}

Evaluation criteria fall into two broad classes, significance tests
and information criteria. This paper considers two significance tests,
the exact conditional test \cite{Kreiner87} and the Log--likelihood
ratio statistic $G^2$ \cite{BishopFH75}, and two
information criteria, Akaike's Information Criterion (AIC)
\cite{Akaike74} and the Bayesian Information Criterion (BIC)
\cite{Schwarz78}. 

\subsection{Significance tests}

The Log-likelihood ratio statistic $G^2$ is defined as:
\begin{equation} G^2 = \sum_{i=1}^q f_i \times log \frac{e_i}{f_i}
\end{equation} where $f_i$ and $e_i$ are the observed and expected
counts of the $i^{th}$ feature vector, respectively.  The observed
count $f_i$ is simply the frequency in the training sample.  The
expected count $e_i$ is calculated from the frequencies in the
training data assuming that the hypothesized model, i.e., the model
generated in the search, adequately fits the sample.  The smaller the
value of $G^2$ the better the fit of the hypothesized model.

The distribution of $G^2$ is asymptotically approximated by the
$\chi^2$ distribution $(G^2 \sim \chi^2$) with adjusted degrees of
freedom (dof) equal to the number of model parameters that have
non-zero estimates given the training sample.  The significance of a
model is equal to the probability of observing its reference $G^2$ in
the $\chi^2$ distribution with appropriate dof.  A hypothesized model
is accepted if the significance (i.e., probability) of its reference
$G^2$ value is greater than, in the case of FSS, or less than, in the
case of BSS, some pre--determined cutoff, $\alpha$.

An alternative to using a $\chi^2$ approximation is to define the
exact conditional distribution of $G^2$.  The exact conditional
distribution of $G^2$ is the distribution of $G^2$ values that would
be observed for comparable data samples randomly generated from the
model being tested.  The significance of $G^2$ based on the exact
conditional distribution does not rely on an asymptotic approximation
and is accurate for sparse and skewed data samples \cite{PedersenKB96}. 

\subsection{Information criteria}

The family of model evaluation criteria known as information criteria
have the following expression: \begin{equation} IC_\kappa = G^2 -
\kappa \times dof \end{equation} where $G^2$ and $dof$ are defined
above. Members of this family are distinguished by their different
values of $\kappa$. AIC corresponds to $\kappa=2$.  BIC corresponds to
$\kappa=log(N)$, where $N$ is the sample size.

The various information criteria are an alternative to using a
pre-defined significance level ($\alpha$) to judge the acceptability
of a model.  AIC and BIC reward good model fit and penalize models
with large numbers of parameters.  The parameter penalty is expressed
as $\kappa \times dof$, where the size of the penalty is the adjusted
degrees of freedom, and the weight of the penalty is controlled by
$\kappa$.

During BSS the hypothesized model with the largest negative
$IC_\kappa$ value is selected as the current model of complexity level
$i-1$, while during FSS the hypothesized model with the largest
positive $IC_\kappa$ value is selected as the current model of
complexity level $i+1$.  The search stops when the $IC_\kappa$ values
for all hypothesized models are greater than zero in the case of BSS,
or less than zero in the case of FSS.


\section{Experimental Data}

The sense--tagged text and feature set used in these experiments
are the same as in \cite{BruceWP96}. The text consists of
every sentence from the ACL/DCI Wall Street Journal corpus 
that contains any of the nouns {\it interest}, {\it bill},
{\it concern}, and {\it drug}, any of the verbs {\it close}, {\it help}, 
{\it agree}, and {\it include}, or any of the adjectives {\it chief}, {\it 
public}, {\it last}, and {\it common}.

The extracted sentences have been hand--tagged with senses defined in
the Longman Dictionary of Contemporary English (LDOCE). There are
between 800 and 3,000 sense--tagged sentences for each of the 12
words. This data was randomly divided into training and test samples
at a 10:1 ratio.

A sentence with an ambiguous word is represented by a feature set with 
three types of contextual feature variables:\footnote{An alternative
feature set for this data is utilized with an exemplar--based learning
algorithm in \cite{NgL96}.} 
(1) The morphological feature
($E$) indicates if an  ambiguous noun is plural or not. For verbs it
indicates  the tense of the verb. This feature is not used for adjectives.  
(2) The POS features have one of 25 possible POS tags, derived from the 
first letter of the tags in the ACL/DCI WSJ corpus. There are four POS 
feature variables representing the POS of the two words immediately 
preceding $(L_1, L_2)$ and following $(R_1, R_2)$ the ambiguous word.  (3) 
The three binary collocation-specific features $(C_1, C_2, C_3)$ indicate 
if  a particular word occurs in a sentence with an ambiguous word. 

The sparse
nature of our data can be illustrated by {\it 
interest}. There are 6 possible values for the sense variable. Combined 
with the other feature variables this  results in 37,500,000 possible 
feature  vectors (or joint parameters). However, we have a training
sample of only 2,100 instances. 

\section{Experimental Results}

In total, eight different decomposable models were selected via a
model search for each of the 12 words.  Each of the eight models
is due to a  different combination of search strategy and
evaluation criterion. Two additional classifiers 
were evaluated to serve as  benchmarks. The 
default classifier  assigns every instance of an ambiguous word
with its  most frequent sense in the training sample. The 
Naive Bayes classifier uses a model that assumes that each contextual
feature variable is conditionally independent of all other contextual
variables  given the value of the sense variable. 

\subsection{Accuracy comparison}

The accuracy\footnote{The percentage of ambiguous words in a held out
test sample that are disambiguated correctly.}   of each of these
classifiers for each of the 12 words  is shown in Figure
\ref{fig:table}. The highest accuracy for each word is in 
bold type while any accuracies less than the default classifier are
italicized. The complexity of the model selected is shown in
parenthesis. For convenience, we refer to  model selection using,
for example, a search strategy of FSS and the evaluation criterion 
AIC as FSS AIC.     

Overall AIC selects the most accurate models during both BSS and
FSS. BSS AIC finds the most accurate model for 6 of 12 words while FSS
AIC finds the most accurate for 4 of 12 words. BSS BIC and the Naive
Bayes find the most accurate model for 3 of 12 words. Each of the other
combinations finds the most most accurate model for 2 of 12 words except
for FSS exact conditional which never finds the most accurate model.  

Neither AIC nor BIC ever selects a model that results in
accuracy less than the default classifier. However, FSS
exact conditional has accuracy less than the default for 6 of 12 words and
BSS exact conditional has accuracy less than the default for 3 of 12 
words. BSS $G^2
\sim \chi^2$ and FSS $G^2 \sim \chi^2$ have less than default accuracy
for 2 of 12 and 1 of 12 words, respectively. 

The accuracy of the significance tests vary greatly depending on the
choice of $\alpha$. Of the various $\alpha$ values that were tested, 
.01, .05, .001, and .0001, the value of .0001 was found 
to produce the most accurate models.
Other values of $\alpha$ will certainly led to other results. 
The information criteria do not require the setting of
any such cut-off values. 

A low complexity model that results in high accuracy
disambiguation is the ultimate goal. Figure \ref{fig:table}
shows that BIC and $G^2 \sim \chi^2$ select lower complexity
models than either AIC or the exact conditional test. However, both 
appear to sacrifice accuracy when compared to AIC. BIC assesses a
greater parameter penalty ($\kappa = log(N)$) than does AIC ($\kappa =
2$), causing BSS BIC to remove more interactions than BSS
AIC. Likewise, FSS BIC adds fewer interactions than FSS AIC. In both
cases BIC selects models whose complexity is too low and adversely
affects accuracy when compared to AIC.    

The Naive Bayes classifier achieves a high level of accuracy using
a model of low complexity. In fact, while the Naive Bayes classifier
is most accurate for only 3 of the 12 words, the average accuracy of
the Naive Bayes classifiers for all 12 words is higher than the
average classification accuracy resulting from any combination of the
search strategies and evaluation criteria. The average complexity of
the Naive Bayes models is also lower than the average complexity of the
models resulting from any combination of the search strategies and
evaluation criteria except BSS BIC and FSS BIC.

\subsection{Search strategy and accuracy}

An evaluation criterion that finds models of similar accuracy using
either BSS or FSS is to be preferred over one that does not. Overall
the information criteria are not greatly affected by a change in the
search strategy, as illustrated in Figure \ref{fig:tstacc}.  Each
point on this plot represents the accuracy of the models selected for a
word by the same evaluation criterion using BSS and FSS. If this point  
falls close to the line $BSS=FSS$ then there is little or no 
difference between the accuracy of the models selected during FSS and BSS. 

AIC exhibits only minor deviation from $BSS=FSS$. This
is also illustrated by the fact that the average accuracy between BSS
AIC and FSS AIC only differs by .0013. The significance tests,
especially the exact conditional,  are more affected by the search
strategy.  It is clear that BSS exact conditional is much more accurate
than FSS exact conditional.   FSS $G^2 \sim \chi^2$ is slightly more
accurate than BSS $G^2 \sim  \chi^2$.  

\setcounter{figure}{2}

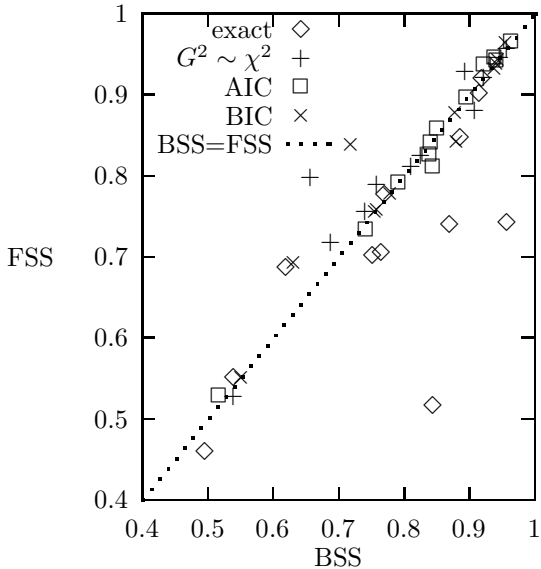
\begin{figure}
\begin{center}
\setlength{\unitlength}{0.240900pt}
\ifx\plotpoint\undefined\newsavebox{\plotpoint}\fi
\sbox{\plotpoint}{\rule[-0.200pt]{0.400pt}{0.400pt}}%
\begin{picture}(900,900)(0,0)
\font\gnuplot=cmr10 at 10pt
\gnuplot
\sbox{\plotpoint}{\rule[-0.200pt]{0.400pt}{0.400pt}}%
\put(220.0,113.0){\rule[-0.200pt]{4.818pt}{0.400pt}}
\put(198,113){\makebox(0,0)[r]{$0.4$}}
\put(816.0,113.0){\rule[-0.200pt]{4.818pt}{0.400pt}}
\put(220.0,240.0){\rule[-0.200pt]{4.818pt}{0.400pt}}
\put(198,240){\makebox(0,0)[r]{$0.5$}}
\put(816.0,240.0){\rule[-0.200pt]{4.818pt}{0.400pt}}
\put(220.0,368.0){\rule[-0.200pt]{4.818pt}{0.400pt}}
\put(198,368){\makebox(0,0)[r]{$0.6$}}
\put(816.0,368.0){\rule[-0.200pt]{4.818pt}{0.400pt}}
\put(220.0,495.0){\rule[-0.200pt]{4.818pt}{0.400pt}}
\put(198,495){\makebox(0,0)[r]{$0.7$}}
\put(816.0,495.0){\rule[-0.200pt]{4.818pt}{0.400pt}}
\put(220.0,622.0){\rule[-0.200pt]{4.818pt}{0.400pt}}
\put(198,622){\makebox(0,0)[r]{$0.8$}}
\put(816.0,622.0){\rule[-0.200pt]{4.818pt}{0.400pt}}
\put(220.0,750.0){\rule[-0.200pt]{4.818pt}{0.400pt}}
\put(198,750){\makebox(0,0)[r]{$0.9$}}
\put(816.0,750.0){\rule[-0.200pt]{4.818pt}{0.400pt}}
\put(220.0,877.0){\rule[-0.200pt]{4.818pt}{0.400pt}}
\put(198,877){\makebox(0,0)[r]{$1$}}
\put(816.0,877.0){\rule[-0.200pt]{4.818pt}{0.400pt}}
\put(220.0,113.0){\rule[-0.200pt]{0.400pt}{4.818pt}}
\put(220,68){\makebox(0,0){$0.4$}}
\put(220.0,857.0){\rule[-0.200pt]{0.400pt}{4.818pt}}
\put(323.0,113.0){\rule[-0.200pt]{0.400pt}{4.818pt}}
\put(323,68){\makebox(0,0){$0.5$}}
\put(323.0,857.0){\rule[-0.200pt]{0.400pt}{4.818pt}}
\put(425.0,113.0){\rule[-0.200pt]{0.400pt}{4.818pt}}
\put(425,68){\makebox(0,0){$0.6$}}
\put(425.0,857.0){\rule[-0.200pt]{0.400pt}{4.818pt}}
\put(528.0,113.0){\rule[-0.200pt]{0.400pt}{4.818pt}}
\put(528,68){\makebox(0,0){$0.7$}}
\put(528.0,857.0){\rule[-0.200pt]{0.400pt}{4.818pt}}
\put(631.0,113.0){\rule[-0.200pt]{0.400pt}{4.818pt}}
\put(631,68){\makebox(0,0){$0.8$}}
\put(631.0,857.0){\rule[-0.200pt]{0.400pt}{4.818pt}}
\put(733.0,113.0){\rule[-0.200pt]{0.400pt}{4.818pt}}
\put(733,68){\makebox(0,0){$0.9$}}
\put(733.0,857.0){\rule[-0.200pt]{0.400pt}{4.818pt}}
\put(836.0,113.0){\rule[-0.200pt]{0.400pt}{4.818pt}}
\put(836,68){\makebox(0,0){$1$}}
\put(836.0,857.0){\rule[-0.200pt]{0.400pt}{4.818pt}}
\put(220.0,113.0){\rule[-0.200pt]{148.394pt}{0.400pt}}
\put(836.0,113.0){\rule[-0.200pt]{0.400pt}{184.048pt}}
\put(220.0,877.0){\rule[-0.200pt]{148.394pt}{0.400pt}}
\put(45,495){\makebox(0,0){FSS}}
\put(528,23){\makebox(0,0){BSS}}
\put(220.0,113.0){\rule[-0.200pt]{0.400pt}{184.048pt}}
\put(425,852){\makebox(0,0)[r]{exact}}
\put(469,852){\raisebox{-.8pt}{\makebox(0,0){$\Diamond$}}}
\put(749,751){\raisebox{-.8pt}{\makebox(0,0){$\Diamond$}}}
\put(445,478){\raisebox{-.8pt}{\makebox(0,0){$\Diamond$}}}
\put(753,775){\raisebox{-.8pt}{\makebox(0,0){$\Diamond$}}}
\put(581,496){\raisebox{-.8pt}{\makebox(0,0){$\Diamond$}}}
\put(702,545){\raisebox{-.8pt}{\makebox(0,0){$\Diamond$}}}
\put(595,501){\raisebox{-.8pt}{\makebox(0,0){$\Diamond$}}}
\put(676,261){\raisebox{-.8pt}{\makebox(0,0){$\Diamond$}}}
\put(600,593){\raisebox{-.8pt}{\makebox(0,0){$\Diamond$}}}
\put(792,549){\raisebox{-.8pt}{\makebox(0,0){$\Diamond$}}}
\put(318,188){\raisebox{-.8pt}{\makebox(0,0){$\Diamond$}}}
\put(719,682){\raisebox{-.8pt}{\makebox(0,0){$\Diamond$}}}
\put(363,305){\raisebox{-.8pt}{\makebox(0,0){$\Diamond$}}}
\sbox{\plotpoint}{\rule[-0.600pt]{1.200pt}{1.200pt}}%
\put(425,807){\makebox(0,0)[r]{$G^2 \sim \chi^2$}}
\put(469,807){\makebox(0,0){$+$}}
\put(727,787){\makebox(0,0){$+$}}
\put(484,620){\makebox(0,0){$+$}}
\put(781,809){\makebox(0,0){$+$}}
\put(588,609){\makebox(0,0){$+$}}
\put(756,777){\makebox(0,0){$+$}}
\put(657,655){\makebox(0,0){$+$}}
\put(642,637){\makebox(0,0){$+$}}
\put(570,566){\makebox(0,0){$+$}}
\put(792,822){\makebox(0,0){$+$}}
\put(516,517){\makebox(0,0){$+$}}
\put(742,725){\makebox(0,0){$+$}}
\put(363,276){\makebox(0,0){$+$}}
\sbox{\plotpoint}{\rule[-0.400pt]{0.800pt}{0.800pt}}%
\put(425,762){\makebox(0,0)[r]{AIC}}
\put(469,762){\raisebox{-.8pt}{\makebox(0,0){$\Box$}}}
\put(756,796){\raisebox{-.8pt}{\makebox(0,0){$\Box$}}}
\put(683,696){\raisebox{-.8pt}{\makebox(0,0){$\Box$}}}
\put(799,832){\raisebox{-.8pt}{\makebox(0,0){$\Box$}}}
\put(673,674){\raisebox{-.8pt}{\makebox(0,0){$\Box$}}}
\put(729,744){\raisebox{-.8pt}{\makebox(0,0){$\Box$}}}
\put(671,655){\raisebox{-.8pt}{\makebox(0,0){$\Box$}}}
\put(676,637){\raisebox{-.8pt}{\makebox(0,0){$\Box$}}}
\put(622,611){\raisebox{-.8pt}{\makebox(0,0){$\Box$}}}
\put(773,807){\raisebox{-.8pt}{\makebox(0,0){$\Box$}}}
\put(571,538){\raisebox{-.8pt}{\makebox(0,0){$\Box$}}}
\put(776,803){\raisebox{-.8pt}{\makebox(0,0){$\Box$}}}
\put(340,276){\raisebox{-.8pt}{\makebox(0,0){$\Box$}}}
\sbox{\plotpoint}{\rule[-0.500pt]{1.000pt}{1.000pt}}%
\put(425,717){\makebox(0,0)[r]{BIC}}
\put(469,717){\makebox(0,0){$\times$}}
\put(778,805){\makebox(0,0){$\times$}}
\put(713,677){\makebox(0,0){$\times$}}
\put(790,832){\makebox(0,0){$\times$}}
\put(588,569){\makebox(0,0){$\times$}}
\put(711,722){\makebox(0,0){$\times$}}
\put(547,672){\makebox(0,0){$\times$}}
\put(609,595){\makebox(0,0){$\times$}}
\put(585,566){\makebox(0,0){$\times$}}
\put(773,791){\makebox(0,0){$\times$}}
\put(457,486){\makebox(0,0){$\times$}}
\put(776,799){\makebox(0,0){$\times$}}
\put(375,305){\makebox(0,0){$\times$}}
\put(425,672){\makebox(0,0)[r]{BSS=FSS}}
\multiput(447,672)(20.756,0.000){4}{\usebox{\plotpoint}}
\put(513,672){\usebox{\plotpoint}}
\put(220,113){\usebox{\plotpoint}}
\put(220.00,113.00){\usebox{\plotpoint}}
\multiput(226,121)(13.508,15.759){0}{\usebox{\plotpoint}}
\put(233.01,129.16){\usebox{\plotpoint}}
\multiput(239,136)(12.453,16.604){0}{\usebox{\plotpoint}}
\put(246.00,145.33){\usebox{\plotpoint}}
\multiput(251,152)(13.508,15.759){0}{\usebox{\plotpoint}}
\put(259.11,161.41){\usebox{\plotpoint}}
\multiput(264,167)(12.453,16.604){0}{\usebox{\plotpoint}}
\put(272.16,177.52){\usebox{\plotpoint}}
\multiput(276,182)(12.453,16.604){0}{\usebox{\plotpoint}}
\put(284.92,193.89){\usebox{\plotpoint}}
\multiput(288,198)(13.668,15.620){0}{\usebox{\plotpoint}}
\put(298.24,209.78){\usebox{\plotpoint}}
\multiput(301,213)(12.453,16.604){0}{\usebox{\plotpoint}}
\put(310.91,226.22){\usebox{\plotpoint}}
\multiput(313,229)(14.676,14.676){0}{\usebox{\plotpoint}}
\put(324.43,241.90){\usebox{\plotpoint}}
\multiput(326,244)(12.453,16.604){0}{\usebox{\plotpoint}}
\put(336.88,258.51){\usebox{\plotpoint}}
\multiput(338,260)(13.508,15.759){0}{\usebox{\plotpoint}}
\put(350.37,274.28){\usebox{\plotpoint}}
\multiput(351,275)(12.453,16.604){0}{\usebox{\plotpoint}}
\multiput(357,283)(13.508,15.759){0}{\usebox{\plotpoint}}
\put(363.34,290.46){\usebox{\plotpoint}}
\multiput(369,298)(13.668,15.620){0}{\usebox{\plotpoint}}
\put(376.42,306.56){\usebox{\plotpoint}}
\multiput(382,314)(13.508,15.759){0}{\usebox{\plotpoint}}
\put(389.34,322.79){\usebox{\plotpoint}}
\multiput(394,329)(12.453,16.604){0}{\usebox{\plotpoint}}
\put(401.97,339.25){\usebox{\plotpoint}}
\multiput(407,345)(13.508,15.759){0}{\usebox{\plotpoint}}
\put(415.34,355.12){\usebox{\plotpoint}}
\multiput(419,360)(12.453,16.604){0}{\usebox{\plotpoint}}
\put(428.29,371.29){\usebox{\plotpoint}}
\multiput(432,375)(12.453,16.604){0}{\usebox{\plotpoint}}
\put(441.30,387.41){\usebox{\plotpoint}}
\multiput(444,391)(12.453,16.604){0}{\usebox{\plotpoint}}
\put(454.08,403.76){\usebox{\plotpoint}}
\multiput(456,406)(13.668,15.620){0}{\usebox{\plotpoint}}
\put(467.30,419.73){\usebox{\plotpoint}}
\multiput(469,422)(13.508,15.759){0}{\usebox{\plotpoint}}
\put(480.22,435.96){\usebox{\plotpoint}}
\multiput(481,437)(13.668,15.620){0}{\usebox{\plotpoint}}
\put(493.30,452.06){\usebox{\plotpoint}}
\multiput(494,453)(13.508,15.759){0}{\usebox{\plotpoint}}
\multiput(500,460)(12.453,16.604){0}{\usebox{\plotpoint}}
\put(506.22,468.29){\usebox{\plotpoint}}
\multiput(512,476)(14.676,14.676){0}{\usebox{\plotpoint}}
\put(519.73,483.98){\usebox{\plotpoint}}
\multiput(525,491)(12.453,16.604){0}{\usebox{\plotpoint}}
\put(532.19,500.58){\usebox{\plotpoint}}
\multiput(537,507)(14.676,14.676){0}{\usebox{\plotpoint}}
\put(545.70,516.27){\usebox{\plotpoint}}
\multiput(550,522)(12.453,16.604){0}{\usebox{\plotpoint}}
\put(558.34,532.72){\usebox{\plotpoint}}
\multiput(562,537)(12.453,16.604){0}{\usebox{\plotpoint}}
\put(571.37,548.86){\usebox{\plotpoint}}
\multiput(575,553)(12.453,16.604){0}{\usebox{\plotpoint}}
\put(584.42,564.99){\usebox{\plotpoint}}
\multiput(587,568)(12.453,16.604){0}{\usebox{\plotpoint}}
\put(597.47,581.11){\usebox{\plotpoint}}
\multiput(600,584)(13.508,15.759){0}{\usebox{\plotpoint}}
\put(610.62,597.15){\usebox{\plotpoint}}
\multiput(612,599)(12.453,16.604){0}{\usebox{\plotpoint}}
\put(623.07,613.76){\usebox{\plotpoint}}
\multiput(624,615)(14.676,14.676){0}{\usebox{\plotpoint}}
\put(636.58,629.44){\usebox{\plotpoint}}
\multiput(637,630)(12.453,16.604){0}{\usebox{\plotpoint}}
\multiput(643,638)(13.508,15.759){0}{\usebox{\plotpoint}}
\put(649.55,645.63){\usebox{\plotpoint}}
\multiput(656,653)(12.453,16.604){0}{\usebox{\plotpoint}}
\put(662.58,661.77){\usebox{\plotpoint}}
\multiput(668,669)(13.508,15.759){0}{\usebox{\plotpoint}}
\put(675.50,678.00){\usebox{\plotpoint}}
\multiput(680,684)(13.668,15.620){0}{\usebox{\plotpoint}}
\put(688.58,694.10){\usebox{\plotpoint}}
\multiput(693,700)(13.508,15.759){0}{\usebox{\plotpoint}}
\put(701.50,710.33){\usebox{\plotpoint}}
\multiput(705,715)(13.668,15.620){0}{\usebox{\plotpoint}}
\put(714.79,726.26){\usebox{\plotpoint}}
\multiput(718,730)(12.453,16.604){0}{\usebox{\plotpoint}}
\put(727.49,742.66){\usebox{\plotpoint}}
\multiput(730,746)(12.453,16.604){0}{\usebox{\plotpoint}}
\put(740.65,758.65){\usebox{\plotpoint}}
\multiput(743,761)(12.453,16.604){0}{\usebox{\plotpoint}}
\put(753.46,774.95){\usebox{\plotpoint}}
\multiput(755,777)(13.508,15.759){0}{\usebox{\plotpoint}}
\put(766.91,790.75){\usebox{\plotpoint}}
\multiput(768,792)(12.453,16.604){0}{\usebox{\plotpoint}}
\put(779.46,807.28){\usebox{\plotpoint}}
\multiput(780,808)(13.508,15.759){0}{\usebox{\plotpoint}}
\multiput(786,815)(12.453,16.604){0}{\usebox{\plotpoint}}
\put(792.42,823.48){\usebox{\plotpoint}}
\multiput(799,831)(13.508,15.759){0}{\usebox{\plotpoint}}
\put(805.92,839.23){\usebox{\plotpoint}}
\multiput(811,846)(12.453,16.604){0}{\usebox{\plotpoint}}
\put(818.51,855.73){\usebox{\plotpoint}}
\multiput(824,862)(13.508,15.759){0}{\usebox{\plotpoint}}
\put(831.92,871.56){\usebox{\plotpoint}}
\put(836,877){\usebox{\plotpoint}}
\end{picture}
\caption{Effect of Search Strategy}
\label{fig:tstacc}
\end{center}
\end{figure}

\subsection{Feature selection: interest}

Figure \ref{fig:models} shows the models selected by the various
combinations of search strategy and evaluation criterion for {\it
interest}. 

During BSS, AIC removed feature $L_2$ from the model,  BIC removed
$L_1, L_2, R_1$ and $R_2$, $G^2 \sim \chi^2$ removed no features, and the 
exact conditional test removed $C_2$. 
During FSS, AIC never added  $R_2$, 
BIC never added $C_1, C_3, L_1, L_2$ and  
$R_2$, and $G^2 \sim \chi^2$ and the exact
conditional test added all the features.  

$G^2 \sim \chi^2$ is the most consistent of the evaluation criteria
in feature selection. During both BSS and FSS it found that all the
features were relevant to classification. 

AIC found seven features to be relevant in both BSS
and FSS. When using AIC, the only difference in the feature
set selected during FSS as compared to that selected during 
BSS is the part of speech feature that is found to be irrelevant: 
during BSS $L_2$ is removed and during FSS $R_2$ is never 
added.  All other criteria exhibit more variation between FSS and 
BSS in feature set selection.

\subsection{Model selection: interest}

Here we consider the results of each stage of the sequential model
selection for {\it interest}.  Figures \ref{fig:bssacc} through
\ref{fig:fssrec} show the accuracy and 
recall\footnote{The
percentage of ambiguous words in a held out test sample that are
disambiguated,  correctly or not. 
A word is not disambiguated if the model parameters needed to assign a
sense tag cannot be estimated from the training sample.}
for the best fitting model at each level of complexity in 
the search. 
The rightmost point on each plot for each evaluation criterion 
is the measure associated with the model ultimately selected.  

These plots illustrate that BSS BIC
selects models of too low complexity. In Figure \ref{fig:bssacc} 
BSS BIC has ``gone past'' much more accurate models than the one it
selected. We observe the related problem for FSS BIC.
In Figure \ref{fig:fssacc} FSS BIC adds too few
interactions and does not select as accurate a model as FSS AIC. 
The exact conditional test suffers from the reverse problem
of BIC. BSS exact conditional removes only a few interactions while
FSS exact conditional adds 
many interactions, and in both cases the resulting models have
poor accuracy.  

The difference between BSS and FSS is clearly illustrated by these plots.
AIC and BIC eliminate interactions that 
have high dof's (and thus have large numbers of parameters) much
earlier in BSS than the significance tests. This rapid reduction in
the number of parameters results in a rapid increases in accuracy
(Figure \ref{fig:bssacc}) and recall for AIC and BIC (Figure
\ref{fig:bssrec}) 
relative to the significance tests as they produce
models with smaller numbers of parameters that can be estimated more
reliably.

However, during the early stages of FSS the number of parameters in the
models is very small and the differences between the information
criteria and the significance tests are minimized. The major
difference among the criteria in Figures \ref{fig:fssacc} and
\ref{fig:fssrec} is that the exact conditional test adds many more
interactions.

\section{Related Work}

Statistical analysis of NLP data has often been limited to the
application of standard models, such as n-gram (Markov chain) models
and the Naive Bayes model. While n-grams perform well in
part--of--speech tagging and speech processing, they require a fixed
interdependency structure that is inappropriate for the broad class of
contextual features used in word--sense disambiguation.  However, the
Naive Bayes classifier has been found to perform well for word--sense
disambiguation both here and in a variety of other works (e.g.,
\cite{BruceW94A}, \cite{GaleCY92}, \cite{LeacockTV93}, and
\cite{Mooney96}).  

In order to utilize models with more complicated interactions among
feature variables, \cite{BruceW94B} introduce the use of sequential
model selection and decomposable models for word--sense
disambiguation.\footnote{They recommended a model selection procedure
using BSS and the exact conditional test in combination with a test
for model predictive power. In their procedure, the exact conditional
test was used to guide the generation of new models and the test of
model predictive power was used to select the final model from among
those generated during the search.}  

Alternative probabilistic approaches have involved using a single
contextual feature to perform disambiguation (e.g., \cite{BrownPPM91},
\cite{DaganIS91}, and \cite{Yarowsky93} present techniques for
identifying the optimal feature to use in disambiguation).  Maximum
Entropy models have been used to express the interactions among
multiple feature variables (e.g., \cite{BergerDD96}), but within this
framework no systematic study of interactions has been proposed.
Decision tree induction has been applied to word-sense disambiguation
(e.g. \cite{Black88} and \cite{Mooney96}) but, while it is a type of model 
selection, the models are not parametric.

\section{Conclusion}

Sequential model selection is a viable means of choosing a probabilistic 
model to perform word--sense disambiguation.  We recommend AIC as 
the evaluation criterion during model selection due to the following: 

\begin{enumerate}

\item It is difficult to set an appropriate cutoff value ($\alpha$) for a 
significance test. 

\item The information criteria AIC and BIC are more robust to changes
in search strategy. 

\item BIC removes too many interactions and results in models of too
low complexity. 
\end{enumerate}

The choice of search strategy when using AIC is less critical than when 
using significance tests. However, we recommend  
FSS for sparse data (NLP data is typically sparse)
since it reduces the impact of very high degrees of freedom
and the resultant unreliable parameter estimates on model selection.

The Naive Bayes classifier is based on a low complexity model that is
shown to lead to high accuracy. If feature selection is not in doubt
(i.e., it is fairly certain that all of the features are somehow
relevant to classification) then this is a reasonable
approach. However, if some features are of questionable value the
Naive Bayes model will continue to utilize them while sequential model
selection will disregard them.  

All of the search strategies and evaluation criteria discussed are
implemented in the public domain program CoCo \cite{Badsberg95}.  


\setcounter{figure}{0}
\onecolumn

\begin{figure} 
\begin{center}
\begin{tabular}{|l|c|c||c|c|c|c|c|} \hline
  & Default & Naive & Search   & $G^2 \sim \chi^2$ & exact & AIC &  BIC
\\
  &         & Bayes &          & $\alpha$ =  .0001 & $\alpha$ = .0001 &    &  \\  \hline
agree & .7660 & .9362 (8)& BSS   &  .8936 (8) &  .9149 (10) &  .9220 (15) & {\bf .9433 (9)} \\ 
 &  &    &            FSS   &  .9291 (12)  & .9007 (15) &  .9362 (13)& {\bf .9433 (7)}  \\ \hline
bill & .7090 & .8657 (8) & BSS & {\it .6567 (22)} &  {\it .6194 (25)}
&  .8507 (26) & {\bf .8806 (7)}  \\ 
      & & & FSS   &               .7985 (20) &  {\it .6866 (28)} &   .8582 (20)
& .8433 (11) \\ \hline
chief & .8750 & {\bf .9643 (7)} & BSS   &  .9464 (6) &  .9196 (17) &
{\bf .9643 (14)} &  .9554  (6) \\
   & & & FSS   &  .9464 (6) &  .9196 (18) &  {\bf .9643 (14)}&{\bf
.9643 (7)} \\ \hline
close & .6815 & .8344 (8) & BSS   &  .7580 (12) &  .7516 (13) &  {\bf .8408 (13)} &  .7580  (3) \\
      & & & FSS   &  .7898 (13)  &  .7006 (19) &  {\bf .8408 (10)} &  .7580 (3)
\\ \hline
common & .8696 & .9130 (7) & BSS  & {\bf  .9217 (4)} &  .8696 (10) &  .8957 (7) &   .8783 (2) \\ 
       & & & FSS   &  {\bf .9217 (4)}  &  {\it .7391 (16)} &   .8957 (7) & .8783 (2)
\\ \hline
concern & .6510 & {\bf .8725 (8)} & BSS & .8255 (5)  &  .7651 (15) &  .8389 (16)
&   .7181 (6) \\
        & & & FSS & .8255 (17)  &  .7047 (24) &   .8255 (13) & .8389  (9)
\\ \hline
drug  & .6721 & .8279 (8) & BSS   & .8115 (10)  &  {\bf .8443 (7)} &
{\bf .8443 (14)} &  .7787  (9) \\ 
      & & & FSS   & .8115 (10)  &  {\it .5164 (19)} &  .8115 (12) &  .7787 (9)
\\ \hline
help & .7266 & .7698 (8) & BSS    & .7410 (7) &  .7698 (6) &  {\bf .7914 (6)} &  .7554 (4)  \\ 
     & & & FSS    & .7554 (3) &  .7770 (9) &  {\bf .7914  (4)}  & .7554 (4) \\
\hline
include & .9325 & .9448 (8) & BSS  & {\bf .9571 (6)} &  {\bf .9571
(3)} & .9387 (16)  &  .9387 (8)  \\ 
        & & & FSS	 & {\bf .9571 (6)}  & {\it .7423 (22)} &  .9448 (9)  & .9325
(9) \\ \hline
interest & .5205 & .7336 (8) & BSS & .6885 (24) &  {\it  .4959 (24)} &
{\bf .7418 (21)} &  .6311 (6)  \\ 
         & & & FSS & .7172 (22)  &  {\it .4590 (32)} &  .7336 (15) &  .6926  (4)
\\ \hline
last  & .9387 & {\it .9264} (7) & BSS    &  {\it .9080 (8)} &  {\it
.8865 (9)} & {\bf .9417 (14)} &  {\bf .9417 (9)}  \\ 
    & & & FSS    &  {\it .8804 (15)}  & {\it .8466 (18)} &  {\bf .9417
(14)} & .9387 (2)
\\ \hline
public & .5056 & {\bf .5843} (7) & BSS  &  .5393 (7) &  .5393 (9) &  .5169 (8) & .5506 (3) \\ 
       & & & FSS  &  .5281 (6)  &  .5506 (11) &  .5281 (6)  & .5506 (3) \\
\hline \hline
average & .7373 & {\bf .8477 (8)} &BSS& .8039 (10)&.7778 (12) &.8406
(14)& .8108 (6)\\  
 & & & FSS  &  .8217 (11)  &  {\it .7119 (19)} &  .8393 (11)&.8229 (6)\\ \hline
\end{tabular}
\caption{Accuracy comparison}
\label{fig:table}
\end{center}
\end{figure}

\begin{figure}
\begin{center}
\begin{tabular}{|l|l|l|} \hline
Criterion & Search & Model \\ \hline
$G^2 \sim \chi^2$ & BSS  & $(C_1 E L_1 L_2 S)(C_1 C_2 C_3 L_1 L_2 S)(C_1
C_2 C_3 R_1 S)$ \\ 
& FSS  & $(C_2 E L_1 L_2 S)(C_1 R_1 R_2 S)(C_2 C_3 L_1 L_2 S)(C_3 R_1 R_2
S)$ \\ \hline
Exact & BSS  & $(C_1 E L_1 L_2 S)(C_1 L_1 L_2 R_1 R_2 S)(C_3 L_1 L_2 R_1
R_2 S)$
 \\
      & FSS  & $(C_1 E L_1 L_2 R_1 R_2 S)(C_3 L_1 L_2 R_1 R_2 S)(C_2 E L_1
L_2 R
_1 R_2 S)$ \\ \hline
AIC   & BSS  & $(C_1 C_2 C_3 E L_1 S)(C_1 C_3 R_1 S)(C_1 C_3 R_2 S)$ \\ 
      & FSS  & $(E L_1 L_2 S)(C_2 E L_2 S)(C_1 R_1 S)(C_3 L_1 S)(C_3 R_1
S) $ \\
 \hline
BIC   & BSS  & $(C_2 E S)(C_1 C_3 S)$ \\ 
      & FSS  & $(C_2 E S)(R_1 S)$ \\ \hline
Naive Bayes & none & $(C_1 S)(C_2 S)(C_3 S)(E S)(L_1 S)(L_2 S)(R_1 S)(R_2
S)$ \\
 \hline
\end{tabular}
\caption{Models selected:  interest}
\label{fig:models}
\end{center}
\end{figure}

\setcounter{figure}{3}
\twocolumn

\begin{figure}
\begin{center}
\setlength{\unitlength}{0.240900pt}
\ifx\plotpoint\undefined\newsavebox{\plotpoint}\fi
\sbox{\plotpoint}{\rule[-0.200pt]{0.400pt}{0.400pt}}%
\begin{picture}(900,900)(0,0)
\font\gnuplot=cmr10 at 10pt
\gnuplot
\sbox{\plotpoint}{\rule[-0.200pt]{0.400pt}{0.400pt}}%
\put(220.0,113.0){\rule[-0.200pt]{4.818pt}{0.400pt}}
\put(198,113){\makebox(0,0)[r]{$0.3$}}
\put(816.0,113.0){\rule[-0.200pt]{4.818pt}{0.400pt}}
\put(220.0,215.0){\rule[-0.200pt]{4.818pt}{0.400pt}}
\put(198,215){\makebox(0,0)[r]{$0.4$}}
\put(816.0,215.0){\rule[-0.200pt]{4.818pt}{0.400pt}}
\put(220.0,317.0){\rule[-0.200pt]{4.818pt}{0.400pt}}
\put(198,317){\makebox(0,0)[r]{$0.5$}}
\put(816.0,317.0){\rule[-0.200pt]{4.818pt}{0.400pt}}
\put(220.0,419.0){\rule[-0.200pt]{4.818pt}{0.400pt}}
\put(198,419){\makebox(0,0)[r]{$0.6$}}
\put(816.0,419.0){\rule[-0.200pt]{4.818pt}{0.400pt}}
\put(220.0,520.0){\rule[-0.200pt]{4.818pt}{0.400pt}}
\put(198,520){\makebox(0,0)[r]{$0.7$}}
\put(816.0,520.0){\rule[-0.200pt]{4.818pt}{0.400pt}}
\put(220.0,622.0){\rule[-0.200pt]{4.818pt}{0.400pt}}
\put(198,622){\makebox(0,0)[r]{$0.8$}}
\put(816.0,622.0){\rule[-0.200pt]{4.818pt}{0.400pt}}
\put(220.0,724.0){\rule[-0.200pt]{4.818pt}{0.400pt}}
\put(198,724){\makebox(0,0)[r]{$0.9$}}
\put(816.0,724.0){\rule[-0.200pt]{4.818pt}{0.400pt}}
\put(220.0,826.0){\rule[-0.200pt]{4.818pt}{0.400pt}}
\put(198,826){\makebox(0,0)[r]{$1$}}
\put(816.0,826.0){\rule[-0.200pt]{4.818pt}{0.400pt}}
\put(836.0,113.0){\rule[-0.200pt]{0.400pt}{4.818pt}}
\put(836,68){\makebox(0,0){$0$}}
\put(836.0,857.0){\rule[-0.200pt]{0.400pt}{4.818pt}}
\put(750.0,113.0){\rule[-0.200pt]{0.400pt}{4.818pt}}
\put(750,68){\makebox(0,0){$5$}}
\put(750.0,857.0){\rule[-0.200pt]{0.400pt}{4.818pt}}
\put(665.0,113.0){\rule[-0.200pt]{0.400pt}{4.818pt}}
\put(665,68){\makebox(0,0){$10$}}
\put(665.0,857.0){\rule[-0.200pt]{0.400pt}{4.818pt}}
\put(579.0,113.0){\rule[-0.200pt]{0.400pt}{4.818pt}}
\put(579,68){\makebox(0,0){$15$}}
\put(579.0,857.0){\rule[-0.200pt]{0.400pt}{4.818pt}}
\put(494.0,113.0){\rule[-0.200pt]{0.400pt}{4.818pt}}
\put(494,68){\makebox(0,0){$20$}}
\put(494.0,857.0){\rule[-0.200pt]{0.400pt}{4.818pt}}
\put(408.0,113.0){\rule[-0.200pt]{0.400pt}{4.818pt}}
\put(408,68){\makebox(0,0){$25$}}
\put(408.0,857.0){\rule[-0.200pt]{0.400pt}{4.818pt}}
\put(323.0,113.0){\rule[-0.200pt]{0.400pt}{4.818pt}}
\put(323,68){\makebox(0,0){$30$}}
\put(323.0,857.0){\rule[-0.200pt]{0.400pt}{4.818pt}}
\put(237.0,113.0){\rule[-0.200pt]{0.400pt}{4.818pt}}
\put(237,68){\makebox(0,0){$35$}}
\put(237.0,857.0){\rule[-0.200pt]{0.400pt}{4.818pt}}
\put(220.0,113.0){\rule[-0.200pt]{148.394pt}{0.400pt}}
\put(836.0,113.0){\rule[-0.200pt]{0.400pt}{184.048pt}}
\put(220.0,877.0){\rule[-0.200pt]{148.394pt}{0.400pt}}
\put(45,495){\makebox(0,0){\%}}
\put(528,23){\makebox(0,0){\# of interactions in model}}
\put(220.0,113.0){\rule[-0.200pt]{0.400pt}{184.048pt}}
\sbox{\plotpoint}{\rule[-0.400pt]{0.800pt}{0.800pt}}%
\put(665,317){\makebox(0,0)[r]{AIC}}
\put(687.0,317.0){\rule[-0.400pt]{15.899pt}{0.800pt}}
\put(477,563){\usebox{\plotpoint}}
\put(460,559.34){\rule{3.600pt}{0.800pt}}
\multiput(469.53,561.34)(-9.528,-4.000){2}{\rule{1.800pt}{0.800pt}}
\multiput(454.57,557.08)(-0.698,-0.509){19}{\rule{1.308pt}{0.123pt}}
\multiput(457.29,557.34)(-15.286,-13.000){2}{\rule{0.654pt}{0.800pt}}
\put(357,542.34){\rule{17.200pt}{0.800pt}}
\multiput(406.30,544.34)(-49.301,-4.000){2}{\rule{8.600pt}{0.800pt}}
\multiput(351.47,540.08)(-0.717,-0.511){17}{\rule{1.333pt}{0.123pt}}
\multiput(354.23,540.34)(-14.233,-12.000){2}{\rule{0.667pt}{0.800pt}}
\multiput(338.09,523.31)(-0.507,-0.897){27}{\rule{0.122pt}{1.612pt}}
\multiput(338.34,526.65)(-17.000,-26.655){2}{\rule{0.800pt}{0.806pt}}
\multiput(315.11,498.08)(-1.139,-0.520){9}{\rule{1.900pt}{0.125pt}}
\multiput(319.06,498.34)(-13.056,-8.000){2}{\rule{0.950pt}{0.800pt}}
\multiput(301.65,490.09)(-0.525,-0.507){27}{\rule{1.047pt}{0.122pt}}
\multiput(303.83,490.34)(-15.827,-17.000){2}{\rule{0.524pt}{0.800pt}}
\multiput(286.09,458.74)(-0.507,-2.413){27}{\rule{0.122pt}{3.918pt}}
\multiput(286.34,466.87)(-17.000,-70.869){2}{\rule{0.800pt}{1.959pt}}
\multiput(263.11,394.08)(-1.139,-0.520){9}{\rule{1.900pt}{0.125pt}}
\multiput(267.06,394.34)(-13.056,-8.000){2}{\rule{0.950pt}{0.800pt}}
\multiput(252.09,370.76)(-0.507,-2.568){27}{\rule{0.122pt}{4.153pt}}
\multiput(252.34,379.38)(-17.000,-75.380){2}{\rule{0.800pt}{2.076pt}}
\multiput(235.09,290.08)(-0.507,-2.042){27}{\rule{0.122pt}{3.353pt}}
\multiput(235.34,297.04)(-17.000,-60.041){2}{\rule{0.800pt}{1.676pt}}
\put(709,317){\makebox(0,0){$\star$}}
\put(477,563){\makebox(0,0){$\star$}}
\put(460,559){\makebox(0,0){$\star$}}
\put(442,546){\makebox(0,0){$\star$}}
\put(357,542){\makebox(0,0){$\star$}}
\put(340,530){\makebox(0,0){$\star$}}
\put(323,500){\makebox(0,0){$\star$}}
\put(306,492){\makebox(0,0){$\star$}}
\put(288,475){\makebox(0,0){$\star$}}
\put(271,396){\makebox(0,0){$\star$}}
\put(254,388){\makebox(0,0){$\star$}}
\put(237,304){\makebox(0,0){$\star$}}
\put(220,237){\makebox(0,0){$\star$}}
\sbox{\plotpoint}{\rule[-0.200pt]{0.400pt}{0.400pt}}%
\put(665,272){\makebox(0,0)[r]{BIC}}
\put(687.0,272.0){\rule[-0.200pt]{15.899pt}{0.400pt}}
\put(733,450){\usebox{\plotpoint}}
\multiput(663.92,450.00)(-0.499,0.535){169}{\rule{0.120pt}{0.528pt}}
\multiput(664.17,450.00)(-86.000,90.904){2}{\rule{0.400pt}{0.264pt}}
\multiput(567.73,542.58)(-3.351,0.493){23}{\rule{2.715pt}{0.119pt}}
\multiput(573.36,541.17)(-79.364,13.000){2}{\rule{1.358pt}{0.400pt}}
\multiput(488.82,553.92)(-1.443,-0.497){57}{\rule{1.247pt}{0.120pt}}
\multiput(491.41,554.17)(-83.412,-30.000){2}{\rule{0.623pt}{0.400pt}}
\multiput(401.94,523.92)(-1.716,-0.497){47}{\rule{1.460pt}{0.120pt}}
\multiput(404.97,524.17)(-81.970,-25.000){2}{\rule{0.730pt}{0.400pt}}
\multiput(320.23,498.92)(-0.712,-0.492){21}{\rule{0.667pt}{0.119pt}}
\multiput(321.62,499.17)(-15.616,-12.000){2}{\rule{0.333pt}{0.400pt}}
\multiput(301.85,486.93)(-1.154,-0.488){13}{\rule{1.000pt}{0.117pt}}
\multiput(303.92,487.17)(-15.924,-8.000){2}{\rule{0.500pt}{0.400pt}}
\multiput(286.92,472.65)(-0.495,-2.122){31}{\rule{0.119pt}{1.771pt}}
\multiput(287.17,476.33)(-17.000,-67.325){2}{\rule{0.400pt}{0.885pt}}
\multiput(269.92,406.53)(-0.495,-0.618){31}{\rule{0.119pt}{0.594pt}}
\multiput(270.17,407.77)(-17.000,-19.767){2}{\rule{0.400pt}{0.297pt}}
\multiput(252.92,379.38)(-0.495,-2.513){31}{\rule{0.119pt}{2.076pt}}
\multiput(253.17,383.69)(-17.000,-79.690){2}{\rule{0.400pt}{1.038pt}}
\multiput(235.92,297.04)(-0.495,-2.001){31}{\rule{0.119pt}{1.676pt}}
\multiput(236.17,300.52)(-17.000,-63.520){2}{\rule{0.400pt}{0.838pt}}
\put(709,272){\raisebox{-.8pt}{\makebox(0,0){$\Diamond$}}}
\put(733,450){\raisebox{-.8pt}{\makebox(0,0){$\Diamond$}}}
\put(716,450){\raisebox{-.8pt}{\makebox(0,0){$\Diamond$}}}
\put(699,450){\raisebox{-.8pt}{\makebox(0,0){$\Diamond$}}}
\put(682,450){\raisebox{-.8pt}{\makebox(0,0){$\Diamond$}}}
\put(665,450){\raisebox{-.8pt}{\makebox(0,0){$\Diamond$}}}
\put(579,542){\raisebox{-.8pt}{\makebox(0,0){$\Diamond$}}}
\put(494,555){\raisebox{-.8pt}{\makebox(0,0){$\Diamond$}}}
\put(408,525){\raisebox{-.8pt}{\makebox(0,0){$\Diamond$}}}
\put(323,500){\raisebox{-.8pt}{\makebox(0,0){$\Diamond$}}}
\put(306,488){\raisebox{-.8pt}{\makebox(0,0){$\Diamond$}}}
\put(288,480){\raisebox{-.8pt}{\makebox(0,0){$\Diamond$}}}
\put(271,409){\raisebox{-.8pt}{\makebox(0,0){$\Diamond$}}}
\put(254,388){\raisebox{-.8pt}{\makebox(0,0){$\Diamond$}}}
\put(237,304){\raisebox{-.8pt}{\makebox(0,0){$\Diamond$}}}
\put(220,237){\raisebox{-.8pt}{\makebox(0,0){$\Diamond$}}}
\put(665.0,450.0){\rule[-0.200pt]{16.381pt}{0.400pt}}
\sbox{\plotpoint}{\rule[-0.500pt]{1.000pt}{1.000pt}}%
\put(665,227){\makebox(0,0)[r]{Exact $\alpha =.0001$}}
\multiput(687,227)(41.511,0.000){2}{\usebox{\plotpoint}}
\put(753,227){\usebox{\plotpoint}}
\put(425,313){\usebox{\plotpoint}}
\multiput(425,313)(-39.381,-13.127){3}{\usebox{\plotpoint}}
\put(310.97,266.97){\usebox{\plotpoint}}
\multiput(306,262)(-31.026,-27.578){0}{\usebox{\plotpoint}}
\put(279.74,239.68){\usebox{\plotpoint}}
\multiput(271,233)(-40.408,-9.508){0}{\usebox{\plotpoint}}
\put(241.30,231.99){\usebox{\plotpoint}}
\multiput(237,233)(40.408,-9.508){0}{\usebox{\plotpoint}}
\put(234.89,233.50){\usebox{\plotpoint}}
\put(220,237){\usebox{\plotpoint}}
\put(709,227){\makebox(0,0){$\triangle$}}
\put(425,313){\makebox(0,0){$\triangle$}}
\put(323,279){\makebox(0,0){$\triangle$}}
\put(306,262){\makebox(0,0){$\triangle$}}
\put(288,246){\makebox(0,0){$\triangle$}}
\put(271,233){\makebox(0,0){$\triangle$}}
\put(254,229){\makebox(0,0){$\triangle$}}
\put(237,233){\makebox(0,0){$\triangle$}}
\put(254,229){\makebox(0,0){$\triangle$}}
\put(220,237){\makebox(0,0){$\triangle$}}
\put(665,182){\makebox(0,0)[r]{$G^2 \sim \chi^2$ $\alpha =.0001$}}
\multiput(687,182)(20.756,0.000){4}{\usebox{\plotpoint}}
\put(753,182){\usebox{\plotpoint}}
\put(425,509){\usebox{\plotpoint}}
\put(425.00,509.00){\usebox{\plotpoint}}
\put(404.80,512.25){\usebox{\plotpoint}}
\multiput(391,509)(-19.709,-6.506){5}{\usebox{\plotpoint}}
\multiput(288,475)(-7.195,-19.469){3}{\usebox{\plotpoint}}
\multiput(271,429)(-6.233,-19.798){2}{\usebox{\plotpoint}}
\multiput(254,375)(-9.505,-18.451){2}{\usebox{\plotpoint}}
\multiput(237,342)(-3.317,-20.489){5}{\usebox{\plotpoint}}
\put(220,237){\usebox{\plotpoint}}
\put(709,182){\circle{12}}
\put(425,509){\circle{12}}
\put(408,513){\circle{12}}
\put(391,509){\circle{12}}
\put(288,475){\circle{12}}
\put(271,429){\circle{12}}
\put(254,375){\circle{12}}
\put(237,342){\circle{12}}
\put(220,237){\circle{12}}
\end{picture}
\caption{BSS accuracy: interest}
\label{fig:bssacc}
\end{center}
\end{figure}

\begin{figure}
\begin{center}
\setlength{\unitlength}{0.240900pt}
\ifx\plotpoint\undefined\newsavebox{\plotpoint}\fi
\sbox{\plotpoint}{\rule[-0.200pt]{0.400pt}{0.400pt}}%
\begin{picture}(900,900)(0,0)
\font\gnuplot=cmr10 at 10pt
\gnuplot
\sbox{\plotpoint}{\rule[-0.200pt]{0.400pt}{0.400pt}}%
\put(220.0,113.0){\rule[-0.200pt]{4.818pt}{0.400pt}}
\put(198,113){\makebox(0,0)[r]{$0.3$}}
\put(816.0,113.0){\rule[-0.200pt]{4.818pt}{0.400pt}}
\put(220.0,215.0){\rule[-0.200pt]{4.818pt}{0.400pt}}
\put(198,215){\makebox(0,0)[r]{$0.4$}}
\put(816.0,215.0){\rule[-0.200pt]{4.818pt}{0.400pt}}
\put(220.0,317.0){\rule[-0.200pt]{4.818pt}{0.400pt}}
\put(198,317){\makebox(0,0)[r]{$0.5$}}
\put(816.0,317.0){\rule[-0.200pt]{4.818pt}{0.400pt}}
\put(220.0,419.0){\rule[-0.200pt]{4.818pt}{0.400pt}}
\put(198,419){\makebox(0,0)[r]{$0.6$}}
\put(816.0,419.0){\rule[-0.200pt]{4.818pt}{0.400pt}}
\put(220.0,520.0){\rule[-0.200pt]{4.818pt}{0.400pt}}
\put(198,520){\makebox(0,0)[r]{$0.7$}}
\put(816.0,520.0){\rule[-0.200pt]{4.818pt}{0.400pt}}
\put(220.0,622.0){\rule[-0.200pt]{4.818pt}{0.400pt}}
\put(198,622){\makebox(0,0)[r]{$0.8$}}
\put(816.0,622.0){\rule[-0.200pt]{4.818pt}{0.400pt}}
\put(220.0,724.0){\rule[-0.200pt]{4.818pt}{0.400pt}}
\put(198,724){\makebox(0,0)[r]{$0.9$}}
\put(816.0,724.0){\rule[-0.200pt]{4.818pt}{0.400pt}}
\put(220.0,826.0){\rule[-0.200pt]{4.818pt}{0.400pt}}
\put(198,826){\makebox(0,0)[r]{$1$}}
\put(816.0,826.0){\rule[-0.200pt]{4.818pt}{0.400pt}}
\put(836.0,113.0){\rule[-0.200pt]{0.400pt}{4.818pt}}
\put(836,68){\makebox(0,0){$0$}}
\put(836.0,857.0){\rule[-0.200pt]{0.400pt}{4.818pt}}
\put(750.0,113.0){\rule[-0.200pt]{0.400pt}{4.818pt}}
\put(750,68){\makebox(0,0){$5$}}
\put(750.0,857.0){\rule[-0.200pt]{0.400pt}{4.818pt}}
\put(665.0,113.0){\rule[-0.200pt]{0.400pt}{4.818pt}}
\put(665,68){\makebox(0,0){$10$}}
\put(665.0,857.0){\rule[-0.200pt]{0.400pt}{4.818pt}}
\put(579.0,113.0){\rule[-0.200pt]{0.400pt}{4.818pt}}
\put(579,68){\makebox(0,0){$15$}}
\put(579.0,857.0){\rule[-0.200pt]{0.400pt}{4.818pt}}
\put(494.0,113.0){\rule[-0.200pt]{0.400pt}{4.818pt}}
\put(494,68){\makebox(0,0){$20$}}
\put(494.0,857.0){\rule[-0.200pt]{0.400pt}{4.818pt}}
\put(408.0,113.0){\rule[-0.200pt]{0.400pt}{4.818pt}}
\put(408,68){\makebox(0,0){$25$}}
\put(408.0,857.0){\rule[-0.200pt]{0.400pt}{4.818pt}}
\put(323.0,113.0){\rule[-0.200pt]{0.400pt}{4.818pt}}
\put(323,68){\makebox(0,0){$30$}}
\put(323.0,857.0){\rule[-0.200pt]{0.400pt}{4.818pt}}
\put(237.0,113.0){\rule[-0.200pt]{0.400pt}{4.818pt}}
\put(237,68){\makebox(0,0){$35$}}
\put(237.0,857.0){\rule[-0.200pt]{0.400pt}{4.818pt}}
\put(220.0,113.0){\rule[-0.200pt]{148.394pt}{0.400pt}}
\put(836.0,113.0){\rule[-0.200pt]{0.400pt}{184.048pt}}
\put(220.0,877.0){\rule[-0.200pt]{148.394pt}{0.400pt}}
\put(45,495){\makebox(0,0){\%}}
\put(528,23){\makebox(0,0){\# of interactions in model}}
\put(220.0,113.0){\rule[-0.200pt]{0.400pt}{184.048pt}}
\sbox{\plotpoint}{\rule[-0.400pt]{0.800pt}{0.800pt}}%
\put(665,317){\makebox(0,0)[r]{AIC}}
\put(687.0,317.0){\rule[-0.400pt]{15.899pt}{0.800pt}}
\put(477,814){\usebox{\plotpoint}}
\multiput(469.90,812.08)(-0.990,-0.516){11}{\rule{1.711pt}{0.124pt}}
\multiput(473.45,812.34)(-13.449,-9.000){2}{\rule{0.856pt}{0.800pt}}
\multiput(424.57,803.09)(-2.599,-0.507){27}{\rule{4.200pt}{0.122pt}}
\multiput(433.28,803.34)(-76.283,-17.000){2}{\rule{2.100pt}{0.800pt}}
\put(340,784.34){\rule{3.600pt}{0.800pt}}
\multiput(349.53,786.34)(-9.528,-4.000){2}{\rule{1.800pt}{0.800pt}}
\put(323,780.34){\rule{3.600pt}{0.800pt}}
\multiput(332.53,782.34)(-9.528,-4.000){2}{\rule{1.800pt}{0.800pt}}
\multiput(321.09,770.18)(-0.507,-1.392){27}{\rule{0.122pt}{2.365pt}}
\multiput(321.34,775.09)(-17.000,-41.092){2}{\rule{0.800pt}{1.182pt}}
\multiput(304.09,720.81)(-0.506,-1.922){29}{\rule{0.122pt}{3.178pt}}
\multiput(304.34,727.40)(-18.000,-60.404){2}{\rule{0.800pt}{1.589pt}}
\multiput(286.09,640.97)(-0.507,-3.960){27}{\rule{0.122pt}{6.271pt}}
\multiput(286.34,653.99)(-17.000,-115.985){2}{\rule{0.800pt}{3.135pt}}
\multiput(269.09,533.07)(-0.507,-0.618){27}{\rule{0.122pt}{1.188pt}}
\multiput(269.34,535.53)(-17.000,-18.534){2}{\rule{0.800pt}{0.594pt}}
\multiput(252.09,497.42)(-0.507,-2.939){27}{\rule{0.122pt}{4.718pt}}
\multiput(252.34,507.21)(-17.000,-86.208){2}{\rule{0.800pt}{2.359pt}}
\multiput(235.09,400.64)(-0.507,-3.063){27}{\rule{0.122pt}{4.906pt}}
\multiput(235.34,410.82)(-17.000,-89.818){2}{\rule{0.800pt}{2.453pt}}
\put(709,317){\makebox(0,0){$\star$}}
\put(477,814){\makebox(0,0){$\star$}}
\put(460,805){\makebox(0,0){$\star$}}
\put(442,805){\makebox(0,0){$\star$}}
\put(357,788){\makebox(0,0){$\star$}}
\put(340,784){\makebox(0,0){$\star$}}
\put(323,780){\makebox(0,0){$\star$}}
\put(306,734){\makebox(0,0){$\star$}}
\put(288,667){\makebox(0,0){$\star$}}
\put(271,538){\makebox(0,0){$\star$}}
\put(254,517){\makebox(0,0){$\star$}}
\put(237,421){\makebox(0,0){$\star$}}
\put(220,321){\makebox(0,0){$\star$}}
\put(442.0,805.0){\rule[-0.400pt]{4.336pt}{0.800pt}}
\sbox{\plotpoint}{\rule[-0.200pt]{0.400pt}{0.400pt}}%
\put(665,272){\makebox(0,0)[r]{BIC}}
\put(687.0,272.0){\rule[-0.200pt]{15.899pt}{0.400pt}}
\put(733,826){\usebox{\plotpoint}}
\multiput(652.69,824.92)(-3.684,-0.492){21}{\rule{2.967pt}{0.119pt}}
\multiput(658.84,825.17)(-79.843,-12.000){2}{\rule{1.483pt}{0.400pt}}
\multiput(550.36,812.93)(-9.393,-0.477){7}{\rule{6.900pt}{0.115pt}}
\multiput(564.68,813.17)(-70.679,-5.000){2}{\rule{3.450pt}{0.400pt}}
\multiput(484.66,807.92)(-2.738,-0.494){29}{\rule{2.250pt}{0.119pt}}
\multiput(489.33,808.17)(-81.330,-16.000){2}{\rule{1.125pt}{0.400pt}}
\multiput(396.73,791.92)(-3.351,-0.493){23}{\rule{2.715pt}{0.119pt}}
\multiput(402.36,792.17)(-79.364,-13.000){2}{\rule{1.358pt}{0.400pt}}
\multiput(321.92,772.26)(-0.495,-2.242){31}{\rule{0.119pt}{1.865pt}}
\multiput(322.17,776.13)(-17.000,-71.130){2}{\rule{0.400pt}{0.932pt}}
\multiput(304.92,698.77)(-0.495,-1.774){33}{\rule{0.119pt}{1.500pt}}
\multiput(305.17,701.89)(-18.000,-59.887){2}{\rule{0.400pt}{0.750pt}}
\multiput(286.92,632.21)(-0.495,-2.874){31}{\rule{0.119pt}{2.359pt}}
\multiput(287.17,637.10)(-17.000,-91.104){2}{\rule{0.400pt}{1.179pt}}
\multiput(269.92,542.75)(-0.495,-0.858){31}{\rule{0.119pt}{0.782pt}}
\multiput(270.17,544.38)(-17.000,-27.376){2}{\rule{0.400pt}{0.391pt}}
\multiput(252.92,507.21)(-0.495,-2.874){31}{\rule{0.119pt}{2.359pt}}
\multiput(253.17,512.10)(-17.000,-91.104){2}{\rule{0.400pt}{1.179pt}}
\multiput(235.92,410.82)(-0.495,-2.994){31}{\rule{0.119pt}{2.453pt}}
\multiput(236.17,415.91)(-17.000,-94.909){2}{\rule{0.400pt}{1.226pt}}
\put(709,272){\raisebox{-.8pt}{\makebox(0,0){$\Diamond$}}}
\put(733,826){\raisebox{-.8pt}{\makebox(0,0){$\Diamond$}}}
\put(716,826){\raisebox{-.8pt}{\makebox(0,0){$\Diamond$}}}
\put(699,826){\raisebox{-.8pt}{\makebox(0,0){$\Diamond$}}}
\put(682,826){\raisebox{-.8pt}{\makebox(0,0){$\Diamond$}}}
\put(665,826){\raisebox{-.8pt}{\makebox(0,0){$\Diamond$}}}
\put(579,814){\raisebox{-.8pt}{\makebox(0,0){$\Diamond$}}}
\put(494,809){\raisebox{-.8pt}{\makebox(0,0){$\Diamond$}}}
\put(408,793){\raisebox{-.8pt}{\makebox(0,0){$\Diamond$}}}
\put(323,780){\raisebox{-.8pt}{\makebox(0,0){$\Diamond$}}}
\put(306,705){\raisebox{-.8pt}{\makebox(0,0){$\Diamond$}}}
\put(288,642){\raisebox{-.8pt}{\makebox(0,0){$\Diamond$}}}
\put(271,546){\raisebox{-.8pt}{\makebox(0,0){$\Diamond$}}}
\put(254,517){\raisebox{-.8pt}{\makebox(0,0){$\Diamond$}}}
\put(237,421){\raisebox{-.8pt}{\makebox(0,0){$\Diamond$}}}
\put(220,321){\raisebox{-.8pt}{\makebox(0,0){$\Diamond$}}}
\put(665.0,826.0){\rule[-0.200pt]{16.381pt}{0.400pt}}
\sbox{\plotpoint}{\rule[-0.500pt]{1.000pt}{1.000pt}}%
\put(665,227){\makebox(0,0)[r]{Exact $\alpha =.0001$}}
\multiput(687,227)(41.511,0.000){2}{\usebox{\plotpoint}}
\put(753,227){\usebox{\plotpoint}}
\put(425,413){\usebox{\plotpoint}}
\multiput(425,413)(-38.384,-15.805){3}{\usebox{\plotpoint}}
\put(314.05,359.94){\usebox{\plotpoint}}
\multiput(306,350)(-30.179,-28.502){0}{\usebox{\plotpoint}}
\put(284.77,330.72){\usebox{\plotpoint}}
\multiput(271,321)(-41.511,0.000){0}{\usebox{\plotpoint}}
\put(246.34,321.00){\usebox{\plotpoint}}
\multiput(237,321)(41.511,0.000){0}{\usebox{\plotpoint}}
\put(238.83,321.00){\usebox{\plotpoint}}
\put(220,321){\usebox{\plotpoint}}
\put(709,227){\makebox(0,0){$\triangle$}}
\put(425,413){\makebox(0,0){$\triangle$}}
\put(323,371){\makebox(0,0){$\triangle$}}
\put(306,350){\makebox(0,0){$\triangle$}}
\put(288,333){\makebox(0,0){$\triangle$}}
\put(271,321){\makebox(0,0){$\triangle$}}
\put(254,321){\makebox(0,0){$\triangle$}}
\put(237,321){\makebox(0,0){$\triangle$}}
\put(254,321){\makebox(0,0){$\triangle$}}
\put(220,321){\makebox(0,0){$\triangle$}}
\put(665,182){\makebox(0,0)[r]{$G^2 \sim \chi^2$ $\alpha =.0001$}}
\multiput(687,182)(20.756,0.000){4}{\usebox{\plotpoint}}
\put(753,182){\usebox{\plotpoint}}
\put(425,722){\usebox{\plotpoint}}
\put(425.00,722.00){\usebox{\plotpoint}}
\put(404.68,720.24){\usebox{\plotpoint}}
\multiput(391,713)(-17.476,-11.198){6}{\usebox{\plotpoint}}
\multiput(288,647)(-6.563,-19.690){3}{\usebox{\plotpoint}}
\multiput(271,596)(-3.227,-20.503){5}{\usebox{\plotpoint}}
\multiput(254,488)(-8.476,-18.946){2}{\usebox{\plotpoint}}
\multiput(237,450)(-2.712,-20.578){6}{\usebox{\plotpoint}}
\put(220,321){\usebox{\plotpoint}}
\put(709,182){\circle{12}}
\put(425,722){\circle{12}}
\put(408,722){\circle{12}}
\put(391,713){\circle{12}}
\put(288,647){\circle{12}}
\put(271,596){\circle{12}}
\put(254,488){\circle{12}}
\put(237,450){\circle{12}}
\put(220,321){\circle{12}}
\end{picture}
\caption{BSS recall: interest}
\label{fig:bssrec}
\end{center}
\end{figure}
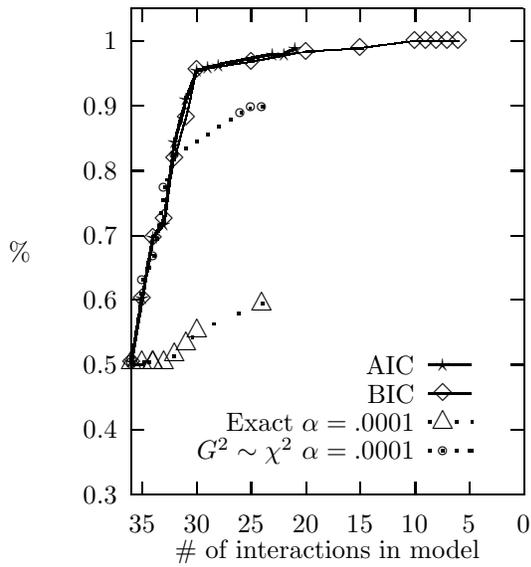

\begin{figure}
\begin{center}
\setlength{\unitlength}{0.240900pt}
\ifx\plotpoint\undefined\newsavebox{\plotpoint}\fi
\sbox{\plotpoint}{\rule[-0.200pt]{0.400pt}{0.400pt}}%
\begin{picture}(900,900)(0,0)
\font\gnuplot=cmr10 at 10pt
\gnuplot
\sbox{\plotpoint}{\rule[-0.200pt]{0.400pt}{0.400pt}}%
\put(220.0,113.0){\rule[-0.200pt]{0.400pt}{184.048pt}}
\put(220.0,113.0){\rule[-0.200pt]{4.818pt}{0.400pt}}
\put(198,113){\makebox(0,0)[r]{$0.3$}}
\put(816.0,113.0){\rule[-0.200pt]{4.818pt}{0.400pt}}
\put(220.0,215.0){\rule[-0.200pt]{4.818pt}{0.400pt}}
\put(198,215){\makebox(0,0)[r]{$0.4$}}
\put(816.0,215.0){\rule[-0.200pt]{4.818pt}{0.400pt}}
\put(220.0,317.0){\rule[-0.200pt]{4.818pt}{0.400pt}}
\put(198,317){\makebox(0,0)[r]{$0.5$}}
\put(816.0,317.0){\rule[-0.200pt]{4.818pt}{0.400pt}}
\put(220.0,419.0){\rule[-0.200pt]{4.818pt}{0.400pt}}
\put(198,419){\makebox(0,0)[r]{$0.6$}}
\put(816.0,419.0){\rule[-0.200pt]{4.818pt}{0.400pt}}
\put(220.0,520.0){\rule[-0.200pt]{4.818pt}{0.400pt}}
\put(198,520){\makebox(0,0)[r]{$0.7$}}
\put(816.0,520.0){\rule[-0.200pt]{4.818pt}{0.400pt}}
\put(220.0,622.0){\rule[-0.200pt]{4.818pt}{0.400pt}}
\put(198,622){\makebox(0,0)[r]{$0.8$}}
\put(816.0,622.0){\rule[-0.200pt]{4.818pt}{0.400pt}}
\put(220.0,724.0){\rule[-0.200pt]{4.818pt}{0.400pt}}
\put(198,724){\makebox(0,0)[r]{$0.9$}}
\put(816.0,724.0){\rule[-0.200pt]{4.818pt}{0.400pt}}
\put(220.0,826.0){\rule[-0.200pt]{4.818pt}{0.400pt}}
\put(198,826){\makebox(0,0)[r]{$1$}}
\put(816.0,826.0){\rule[-0.200pt]{4.818pt}{0.400pt}}
\put(220.0,113.0){\rule[-0.200pt]{0.400pt}{4.818pt}}
\put(220,68){\makebox(0,0){$0$}}
\put(220.0,857.0){\rule[-0.200pt]{0.400pt}{4.818pt}}
\put(306.0,113.0){\rule[-0.200pt]{0.400pt}{4.818pt}}
\put(306,68){\makebox(0,0){$5$}}
\put(306.0,857.0){\rule[-0.200pt]{0.400pt}{4.818pt}}
\put(391.0,113.0){\rule[-0.200pt]{0.400pt}{4.818pt}}
\put(391,68){\makebox(0,0){$10$}}
\put(391.0,857.0){\rule[-0.200pt]{0.400pt}{4.818pt}}
\put(477.0,113.0){\rule[-0.200pt]{0.400pt}{4.818pt}}
\put(477,68){\makebox(0,0){$15$}}
\put(477.0,857.0){\rule[-0.200pt]{0.400pt}{4.818pt}}
\put(562.0,113.0){\rule[-0.200pt]{0.400pt}{4.818pt}}
\put(562,68){\makebox(0,0){$20$}}
\put(562.0,857.0){\rule[-0.200pt]{0.400pt}{4.818pt}}
\put(648.0,113.0){\rule[-0.200pt]{0.400pt}{4.818pt}}
\put(648,68){\makebox(0,0){$25$}}
\put(648.0,857.0){\rule[-0.200pt]{0.400pt}{4.818pt}}
\put(733.0,113.0){\rule[-0.200pt]{0.400pt}{4.818pt}}
\put(733,68){\makebox(0,0){$30$}}
\put(733.0,857.0){\rule[-0.200pt]{0.400pt}{4.818pt}}
\put(819.0,113.0){\rule[-0.200pt]{0.400pt}{4.818pt}}
\put(819,68){\makebox(0,0){$35$}}
\put(819.0,857.0){\rule[-0.200pt]{0.400pt}{4.818pt}}
\put(220.0,113.0){\rule[-0.200pt]{148.394pt}{0.400pt}}
\put(836.0,113.0){\rule[-0.200pt]{0.400pt}{184.048pt}}
\put(220.0,877.0){\rule[-0.200pt]{148.394pt}{0.400pt}}
\put(45,495){\makebox(0,0){\%}}
\put(528,23){\makebox(0,0){\# of interactions in model}}
\put(220.0,113.0){\rule[-0.200pt]{0.400pt}{184.048pt}}
\sbox{\plotpoint}{\rule[-0.400pt]{0.800pt}{0.800pt}}%
\put(596,317){\makebox(0,0)[r]{AIC}}
\put(618.0,317.0){\rule[-0.400pt]{15.899pt}{0.800pt}}
\put(477,555){\usebox{\plotpoint}}
\multiput(469.11,556.40)(-1.139,0.520){9}{\rule{1.900pt}{0.125pt}}
\multiput(473.06,553.34)(-13.056,8.000){2}{\rule{0.950pt}{0.800pt}}
\put(442,559.34){\rule{3.800pt}{0.800pt}}
\multiput(452.11,561.34)(-10.113,-4.000){2}{\rule{1.900pt}{0.800pt}}
\multiput(406.09,552.50)(-0.507,-0.866){27}{\rule{0.122pt}{1.565pt}}
\multiput(406.34,555.75)(-17.000,-25.752){2}{\rule{0.800pt}{0.782pt}}
\put(408.0,559.0){\rule[-0.400pt]{8.191pt}{0.800pt}}
\multiput(351.83,528.08)(-0.657,-0.509){19}{\rule{1.246pt}{0.123pt}}
\multiput(354.41,528.34)(-14.414,-13.000){2}{\rule{0.623pt}{0.800pt}}
\multiput(338.09,505.62)(-0.507,-1.639){27}{\rule{0.122pt}{2.741pt}}
\multiput(338.34,511.31)(-17.000,-48.311){2}{\rule{0.800pt}{1.371pt}}
\multiput(321.09,463.00)(-0.507,1.392){27}{\rule{0.122pt}{2.365pt}}
\multiput(321.34,463.00)(-17.000,41.092){2}{\rule{0.800pt}{1.182pt}}
\multiput(304.09,504.18)(-0.503,-0.600){63}{\rule{0.121pt}{1.160pt}}
\multiput(304.34,506.59)(-35.000,-39.592){2}{\rule{0.800pt}{0.580pt}}
\multiput(265.83,465.08)(-0.657,-0.509){19}{\rule{1.246pt}{0.123pt}}
\multiput(268.41,465.34)(-14.414,-13.000){2}{\rule{0.623pt}{0.800pt}}
\put(357.0,530.0){\rule[-0.400pt]{8.191pt}{0.800pt}}
\multiput(235.09,430.51)(-0.507,-3.558){27}{\rule{0.122pt}{5.659pt}}
\multiput(235.34,442.25)(-17.000,-104.255){2}{\rule{0.800pt}{2.829pt}}
\put(640,317){\makebox(0,0){$\star$}}
\put(477,555){\makebox(0,0){$\star$}}
\put(460,563){\makebox(0,0){$\star$}}
\put(442,559){\makebox(0,0){$\star$}}
\put(408,559){\makebox(0,0){$\star$}}
\put(391,530){\makebox(0,0){$\star$}}
\put(357,530){\makebox(0,0){$\star$}}
\put(340,517){\makebox(0,0){$\star$}}
\put(323,463){\makebox(0,0){$\star$}}
\put(306,509){\makebox(0,0){$\star$}}
\put(271,467){\makebox(0,0){$\star$}}
\put(254,454){\makebox(0,0){$\star$}}
\put(237,454){\makebox(0,0){$\star$}}
\put(220,338){\makebox(0,0){$\star$}}
\put(237.0,454.0){\rule[-0.400pt]{4.095pt}{0.800pt}}
\sbox{\plotpoint}{\rule[-0.200pt]{0.400pt}{0.400pt}}%
\put(596,272){\makebox(0,0)[r]{BIC}}
\put(618.0,272.0){\rule[-0.200pt]{15.899pt}{0.400pt}}
\put(288,513){\usebox{\plotpoint}}
\multiput(280.53,513.60)(-2.382,0.468){5}{\rule{1.800pt}{0.113pt}}
\multiput(284.26,512.17)(-13.264,4.000){2}{\rule{0.900pt}{0.400pt}}
\multiput(269.92,510.43)(-0.495,-1.881){31}{\rule{0.119pt}{1.582pt}}
\multiput(270.17,513.72)(-17.000,-59.716){2}{\rule{0.400pt}{0.791pt}}
\multiput(235.92,442.25)(-0.495,-3.475){31}{\rule{0.119pt}{2.829pt}}
\multiput(236.17,448.13)(-17.000,-110.127){2}{\rule{0.400pt}{1.415pt}}
\put(640,272){\raisebox{-.8pt}{\makebox(0,0){$\Diamond$}}}
\put(288,513){\raisebox{-.8pt}{\makebox(0,0){$\Diamond$}}}
\put(271,517){\raisebox{-.8pt}{\makebox(0,0){$\Diamond$}}}
\put(254,454){\raisebox{-.8pt}{\makebox(0,0){$\Diamond$}}}
\put(237,454){\raisebox{-.8pt}{\makebox(0,0){$\Diamond$}}}
\put(220,338){\raisebox{-.8pt}{\makebox(0,0){$\Diamond$}}}
\put(237.0,454.0){\rule[-0.200pt]{4.095pt}{0.400pt}}
\sbox{\plotpoint}{\rule[-0.500pt]{1.000pt}{1.000pt}}%
\put(596,227){\makebox(0,0)[r]{Exact $\alpha =.0001$}}
\multiput(618,227)(41.511,0.000){2}{\usebox{\plotpoint}}
\put(684,227){\usebox{\plotpoint}}
\put(768,275){\usebox{\plotpoint}}
\put(768.00,275.00){\usebox{\plotpoint}}
\put(737.73,301.04){\usebox{\plotpoint}}
\multiput(733,308)(-39.824,11.713){0}{\usebox{\plotpoint}}
\multiput(716,313)(-10.977,40.033){2}{\usebox{\plotpoint}}
\multiput(699,375)(-26.119,32.264){0}{\usebox{\plotpoint}}
\multiput(682,396)(-40.773,7.795){2}{\usebox{\plotpoint}}
\multiput(614,409)(-35.887,20.864){2}{\usebox{\plotpoint}}
\put(524.39,461.55){\usebox{\plotpoint}}
\multiput(511,471)(-29.353,29.353){0}{\usebox{\plotpoint}}
\put(493.24,488.76){\usebox{\plotpoint}}
\put(460.89,514.20){\usebox{\plotpoint}}
\multiput(442,525)(-41.511,0.000){0}{\usebox{\plotpoint}}
\put(422.25,525.00){\usebox{\plotpoint}}
\put(389.55,504.93){\usebox{\plotpoint}}
\put(357.24,495.89){\usebox{\plotpoint}}
\multiput(357,496)(-41.511,0.000){0}{\usebox{\plotpoint}}
\put(326.37,475.95){\usebox{\plotpoint}}
\put(300.83,447.15){\usebox{\plotpoint}}
\multiput(288,450)(-30.228,-28.450){0}{\usebox{\plotpoint}}
\put(266.11,435.15){\usebox{\plotpoint}}
\multiput(254,438)(-8.234,-40.686){2}{\usebox{\plotpoint}}
\multiput(237,354)(-30.228,-28.450){0}{\usebox{\plotpoint}}
\put(220,338){\usebox{\plotpoint}}
\put(640,227){\makebox(0,0){$\triangle$}}
\put(768,275){\makebox(0,0){$\triangle$}}
\put(750,283){\makebox(0,0){$\triangle$}}
\put(733,308){\makebox(0,0){$\triangle$}}
\put(716,313){\makebox(0,0){$\triangle$}}
\put(699,375){\makebox(0,0){$\triangle$}}
\put(682,396){\makebox(0,0){$\triangle$}}
\put(614,409){\makebox(0,0){$\triangle$}}
\put(528,459){\makebox(0,0){$\triangle$}}
\put(511,471){\makebox(0,0){$\triangle$}}
\put(494,488){\makebox(0,0){$\triangle$}}
\put(477,505){\makebox(0,0){$\triangle$}}
\put(442,525){\makebox(0,0){$\triangle$}}
\put(425,525){\makebox(0,0){$\triangle$}}
\put(408,525){\makebox(0,0){$\triangle$}}
\put(374,488){\makebox(0,0){$\triangle$}}
\put(357,496){\makebox(0,0){$\triangle$}}
\put(340,496){\makebox(0,0){$\triangle$}}
\put(306,446){\makebox(0,0){$\triangle$}}
\put(288,450){\makebox(0,0){$\triangle$}}
\put(271,434){\makebox(0,0){$\triangle$}}
\put(254,438){\makebox(0,0){$\triangle$}}
\put(237,354){\makebox(0,0){$\triangle$}}
\put(220,338){\makebox(0,0){$\triangle$}}
\put(596,182){\makebox(0,0)[r]{$G^2 \sim \chi^2$ $\alpha =.0001$}}
\multiput(618,182)(20.756,0.000){4}{\usebox{\plotpoint}}
\put(684,182){\usebox{\plotpoint}}
\put(596,538){\usebox{\plotpoint}}
\put(596.00,538.00){\usebox{\plotpoint}}
\multiput(579,530)(-13.059,16.132){2}{\usebox{\plotpoint}}
\put(546.21,546.36){\usebox{\plotpoint}}
\multiput(545,546)(-20.756,0.000){0}{\usebox{\plotpoint}}
\put(526.02,547.52){\usebox{\plotpoint}}
\put(509.15,559.00){\usebox{\plotpoint}}
\multiput(494,559)(-20.613,-2.425){2}{\usebox{\plotpoint}}
\multiput(460,555)(-20.756,0.000){2}{\usebox{\plotpoint}}
\multiput(425,555)(-20.204,4.754){0}{\usebox{\plotpoint}}
\multiput(408,559)(-10.496,-17.906){2}{\usebox{\plotpoint}}
\multiput(391,530)(-20.756,0.000){2}{\usebox{\plotpoint}}
\put(343.23,519.47){\usebox{\plotpoint}}
\multiput(340,517)(-6.233,-19.798){2}{\usebox{\plotpoint}}
\multiput(323,463)(-7.195,19.469){3}{\usebox{\plotpoint}}
\multiput(306,509)(-13.287,-15.945){2}{\usebox{\plotpoint}}
\put(269.69,466.00){\usebox{\plotpoint}}
\put(252.99,454.00){\usebox{\plotpoint}}
\multiput(237,454)(-3.010,-20.536){6}{\usebox{\plotpoint}}
\put(220,338){\usebox{\plotpoint}}
\put(640,182){\circle{12}}
\put(596,538){\circle{12}}
\put(579,530){\circle{12}}
\put(562,551){\circle{12}}
\put(545,546){\circle{12}}
\put(528,546){\circle{12}}
\put(511,559){\circle{12}}
\put(494,559){\circle{12}}
\put(460,555){\circle{12}}
\put(425,555){\circle{12}}
\put(408,559){\circle{12}}
\put(391,530){\circle{12}}
\put(357,530){\circle{12}}
\put(340,517){\circle{12}}
\put(323,463){\circle{12}}
\put(306,509){\circle{12}}
\put(271,467){\circle{12}}
\put(254,454){\circle{12}}
\put(237,454){\circle{12}}
\put(220,338){\circle{12}}
\end{picture}
\caption{FSS accuracy: interest}
\label{fig:fssacc}
\end{center}
\end{figure}

\begin{figure}
\begin{center}
\setlength{\unitlength}{0.240900pt}
\ifx\plotpoint\undefined\newsavebox{\plotpoint}\fi
\sbox{\plotpoint}{\rule[-0.200pt]{0.400pt}{0.400pt}}%
\begin{picture}(900,900)(0,0)
\font\gnuplot=cmr10 at 10pt
\gnuplot
\sbox{\plotpoint}{\rule[-0.200pt]{0.400pt}{0.400pt}}%
\put(220.0,113.0){\rule[-0.200pt]{0.400pt}{184.048pt}}
\put(220.0,113.0){\rule[-0.200pt]{4.818pt}{0.400pt}}
\put(198,113){\makebox(0,0)[r]{$0.3$}}
\put(816.0,113.0){\rule[-0.200pt]{4.818pt}{0.400pt}}
\put(220.0,215.0){\rule[-0.200pt]{4.818pt}{0.400pt}}
\put(198,215){\makebox(0,0)[r]{$0.4$}}
\put(816.0,215.0){\rule[-0.200pt]{4.818pt}{0.400pt}}
\put(220.0,317.0){\rule[-0.200pt]{4.818pt}{0.400pt}}
\put(198,317){\makebox(0,0)[r]{$0.5$}}
\put(816.0,317.0){\rule[-0.200pt]{4.818pt}{0.400pt}}
\put(220.0,419.0){\rule[-0.200pt]{4.818pt}{0.400pt}}
\put(198,419){\makebox(0,0)[r]{$0.6$}}
\put(816.0,419.0){\rule[-0.200pt]{4.818pt}{0.400pt}}
\put(220.0,520.0){\rule[-0.200pt]{4.818pt}{0.400pt}}
\put(198,520){\makebox(0,0)[r]{$0.7$}}
\put(816.0,520.0){\rule[-0.200pt]{4.818pt}{0.400pt}}
\put(220.0,622.0){\rule[-0.200pt]{4.818pt}{0.400pt}}
\put(198,622){\makebox(0,0)[r]{$0.8$}}
\put(816.0,622.0){\rule[-0.200pt]{4.818pt}{0.400pt}}
\put(220.0,724.0){\rule[-0.200pt]{4.818pt}{0.400pt}}
\put(198,724){\makebox(0,0)[r]{$0.9$}}
\put(816.0,724.0){\rule[-0.200pt]{4.818pt}{0.400pt}}
\put(220.0,826.0){\rule[-0.200pt]{4.818pt}{0.400pt}}
\put(198,826){\makebox(0,0)[r]{$1$}}
\put(816.0,826.0){\rule[-0.200pt]{4.818pt}{0.400pt}}
\put(220.0,113.0){\rule[-0.200pt]{0.400pt}{4.818pt}}
\put(220,68){\makebox(0,0){$0$}}
\put(220.0,857.0){\rule[-0.200pt]{0.400pt}{4.818pt}}
\put(306.0,113.0){\rule[-0.200pt]{0.400pt}{4.818pt}}
\put(306,68){\makebox(0,0){$5$}}
\put(306.0,857.0){\rule[-0.200pt]{0.400pt}{4.818pt}}
\put(391.0,113.0){\rule[-0.200pt]{0.400pt}{4.818pt}}
\put(391,68){\makebox(0,0){$10$}}
\put(391.0,857.0){\rule[-0.200pt]{0.400pt}{4.818pt}}
\put(477.0,113.0){\rule[-0.200pt]{0.400pt}{4.818pt}}
\put(477,68){\makebox(0,0){$15$}}
\put(477.0,857.0){\rule[-0.200pt]{0.400pt}{4.818pt}}
\put(562.0,113.0){\rule[-0.200pt]{0.400pt}{4.818pt}}
\put(562,68){\makebox(0,0){$20$}}
\put(562.0,857.0){\rule[-0.200pt]{0.400pt}{4.818pt}}
\put(648.0,113.0){\rule[-0.200pt]{0.400pt}{4.818pt}}
\put(648,68){\makebox(0,0){$25$}}
\put(648.0,857.0){\rule[-0.200pt]{0.400pt}{4.818pt}}
\put(733.0,113.0){\rule[-0.200pt]{0.400pt}{4.818pt}}
\put(733,68){\makebox(0,0){$30$}}
\put(733.0,857.0){\rule[-0.200pt]{0.400pt}{4.818pt}}
\put(819.0,113.0){\rule[-0.200pt]{0.400pt}{4.818pt}}
\put(819,68){\makebox(0,0){$35$}}
\put(819.0,857.0){\rule[-0.200pt]{0.400pt}{4.818pt}}
\put(220.0,113.0){\rule[-0.200pt]{148.394pt}{0.400pt}}
\put(836.0,113.0){\rule[-0.200pt]{0.400pt}{184.048pt}}
\put(220.0,877.0){\rule[-0.200pt]{148.394pt}{0.400pt}}
\put(45,495){\makebox(0,0){\%}}
\put(528,23){\makebox(0,0){\# of interactions in model}}
\put(220.0,113.0){\rule[-0.200pt]{0.400pt}{184.048pt}}
\sbox{\plotpoint}{\rule[-0.400pt]{0.800pt}{0.800pt}}%
\put(596,317){\makebox(0,0)[r]{AIC}}
\put(618.0,317.0){\rule[-0.400pt]{15.899pt}{0.800pt}}
\put(477,801){\usebox{\plotpoint}}
\put(391,801.34){\rule{3.600pt}{0.800pt}}
\multiput(400.53,799.34)(-9.528,4.000){2}{\rule{1.800pt}{0.800pt}}
\put(408.0,801.0){\rule[-0.400pt]{16.622pt}{0.800pt}}
\put(323,805.34){\rule{3.600pt}{0.800pt}}
\multiput(332.53,803.34)(-9.528,4.000){2}{\rule{1.800pt}{0.800pt}}
\put(340.0,805.0){\rule[-0.400pt]{12.286pt}{0.800pt}}
\multiput(298.33,810.41)(-1.051,0.507){27}{\rule{1.847pt}{0.122pt}}
\multiput(302.17,807.34)(-31.166,17.000){2}{\rule{0.924pt}{0.800pt}}
\put(306.0,809.0){\rule[-0.400pt]{4.095pt}{0.800pt}}
\put(640,317){\makebox(0,0){$\star$}}
\put(477,801){\makebox(0,0){$\star$}}
\put(460,801){\makebox(0,0){$\star$}}
\put(442,801){\makebox(0,0){$\star$}}
\put(408,801){\makebox(0,0){$\star$}}
\put(391,805){\makebox(0,0){$\star$}}
\put(357,805){\makebox(0,0){$\star$}}
\put(340,805){\makebox(0,0){$\star$}}
\put(323,809){\makebox(0,0){$\star$}}
\put(306,809){\makebox(0,0){$\star$}}
\put(271,826){\makebox(0,0){$\star$}}
\put(254,826){\makebox(0,0){$\star$}}
\put(237,826){\makebox(0,0){$\star$}}
\put(220,826){\makebox(0,0){$\star$}}
\put(220.0,826.0){\rule[-0.400pt]{12.286pt}{0.800pt}}
\sbox{\plotpoint}{\rule[-0.200pt]{0.400pt}{0.400pt}}%
\put(596,272){\makebox(0,0)[r]{BIC}}
\put(618.0,272.0){\rule[-0.200pt]{15.899pt}{0.400pt}}
\put(288,826){\usebox{\plotpoint}}
\put(640,272){\raisebox{-.8pt}{\makebox(0,0){$\Diamond$}}}
\put(288,826){\raisebox{-.8pt}{\makebox(0,0){$\Diamond$}}}
\put(271,826){\raisebox{-.8pt}{\makebox(0,0){$\Diamond$}}}
\put(254,826){\raisebox{-.8pt}{\makebox(0,0){$\Diamond$}}}
\put(237,826){\raisebox{-.8pt}{\makebox(0,0){$\Diamond$}}}
\put(220,826){\raisebox{-.8pt}{\makebox(0,0){$\Diamond$}}}
\put(220.0,826.0){\rule[-0.200pt]{16.381pt}{0.400pt}}
\sbox{\plotpoint}{\rule[-0.500pt]{1.000pt}{1.000pt}}%
\put(596,227){\makebox(0,0)[r]{Exact $\alpha =.0001$}}
\multiput(618,227)(41.511,0.000){2}{\usebox{\plotpoint}}
\put(684,227){\usebox{\plotpoint}}
\put(768,363){\usebox{\plotpoint}}
\put(768.00,363.00){\usebox{\plotpoint}}
\put(734.21,387.07){\usebox{\plotpoint}}
\multiput(733,388)(-30.228,28.450){0}{\usebox{\plotpoint}}
\multiput(716,404)(-10.815,40.078){2}{\usebox{\plotpoint}}
\put(687.07,499.28){\usebox{\plotpoint}}
\multiput(682,513)(-32.508,25.815){2}{\usebox{\plotpoint}}
\multiput(614,567)(-34.971,22.365){2}{\usebox{\plotpoint}}
\put(526.16,625.56){\usebox{\plotpoint}}
\multiput(511,655)(-9.176,40.484){2}{\usebox{\plotpoint}}
\multiput(494,730)(-11.676,39.835){2}{\usebox{\plotpoint}}
\multiput(477,788)(-41.094,5.871){0}{\usebox{\plotpoint}}
\put(440.26,793.00){\usebox{\plotpoint}}
\multiput(425,793)(-41.511,0.000){0}{\usebox{\plotpoint}}
\put(398.74,793.00){\usebox{\plotpoint}}
\put(360.30,802.67){\usebox{\plotpoint}}
\multiput(357,805)(-29.353,29.353){0}{\usebox{\plotpoint}}
\put(326.57,822.00){\usebox{\plotpoint}}
\multiput(306,822)(-41.511,0.000){0}{\usebox{\plotpoint}}
\put(285.06,822.00){\usebox{\plotpoint}}
\multiput(271,822)(-41.511,0.000){0}{\usebox{\plotpoint}}
\put(243.83,824.39){\usebox{\plotpoint}}
\multiput(237,826)(-41.511,0.000){0}{\usebox{\plotpoint}}
\put(220,826){\usebox{\plotpoint}}
\put(640,227){\makebox(0,0){$\triangle$}}
\put(768,363){\makebox(0,0){$\triangle$}}
\put(750,375){\makebox(0,0){$\triangle$}}
\put(733,388){\makebox(0,0){$\triangle$}}
\put(716,404){\makebox(0,0){$\triangle$}}
\put(699,467){\makebox(0,0){$\triangle$}}
\put(682,513){\makebox(0,0){$\triangle$}}
\put(614,567){\makebox(0,0){$\triangle$}}
\put(528,622){\makebox(0,0){$\triangle$}}
\put(511,655){\makebox(0,0){$\triangle$}}
\put(494,730){\makebox(0,0){$\triangle$}}
\put(477,788){\makebox(0,0){$\triangle$}}
\put(442,793){\makebox(0,0){$\triangle$}}
\put(425,793){\makebox(0,0){$\triangle$}}
\put(408,793){\makebox(0,0){$\triangle$}}
\put(374,793){\makebox(0,0){$\triangle$}}
\put(357,805){\makebox(0,0){$\triangle$}}
\put(340,822){\makebox(0,0){$\triangle$}}
\put(306,822){\makebox(0,0){$\triangle$}}
\put(288,822){\makebox(0,0){$\triangle$}}
\put(271,822){\makebox(0,0){$\triangle$}}
\put(254,822){\makebox(0,0){$\triangle$}}
\put(237,826){\makebox(0,0){$\triangle$}}
\put(220,826){\makebox(0,0){$\triangle$}}
\put(596,182){\makebox(0,0)[r]{$G^2 \sim \chi^2$ $\alpha =.0001$}}
\multiput(618,182)(20.756,0.000){4}{\usebox{\plotpoint}}
\put(684,182){\usebox{\plotpoint}}
\put(596,717){\usebox{\plotpoint}}
\put(596.00,717.00){\usebox{\plotpoint}}
\multiput(579,717)(-10.233,18.058){2}{\usebox{\plotpoint}}
\put(551.22,747.00){\usebox{\plotpoint}}
\put(531.84,753.19){\usebox{\plotpoint}}
\put(516.33,766.67){\usebox{\plotpoint}}
\put(497.78,772.00){\usebox{\plotpoint}}
\multiput(494,772)(-16.722,12.295){2}{\usebox{\plotpoint}}
\put(443.82,798.85){\usebox{\plotpoint}}
\put(423.19,801.00){\usebox{\plotpoint}}
\put(402.58,802.28){\usebox{\plotpoint}}
\multiput(391,805)(-20.756,0.000){2}{\usebox{\plotpoint}}
\put(340.63,805.00){\usebox{\plotpoint}}
\multiput(340,805)(-20.204,4.754){0}{\usebox{\plotpoint}}
\put(320.34,809.00){\usebox{\plotpoint}}
\multiput(306,809)(-18.670,9.068){2}{\usebox{\plotpoint}}
\put(261.98,826.00){\usebox{\plotpoint}}
\put(241.23,826.00){\usebox{\plotpoint}}
\put(220.47,826.00){\usebox{\plotpoint}}
\put(220,826){\usebox{\plotpoint}}
\put(640,182){\circle{12}}
\put(596,717){\circle{12}}
\put(579,717){\circle{12}}
\put(562,747){\circle{12}}
\put(545,747){\circle{12}}
\put(528,755){\circle{12}}
\put(511,772){\circle{12}}
\put(494,772){\circle{12}}
\put(460,797){\circle{12}}
\put(425,801){\circle{12}}
\put(408,801){\circle{12}}
\put(391,805){\circle{12}}
\put(357,805){\circle{12}}
\put(340,805){\circle{12}}
\put(323,809){\circle{12}}
\put(306,809){\circle{12}}
\put(271,826){\circle{12}}
\put(254,826){\circle{12}}
\put(237,826){\circle{12}}
\put(220,826){\circle{12}}
\end{picture}
\caption{FSS recall: interest}
\label{fig:fssrec}
\end{center}
\end{figure}
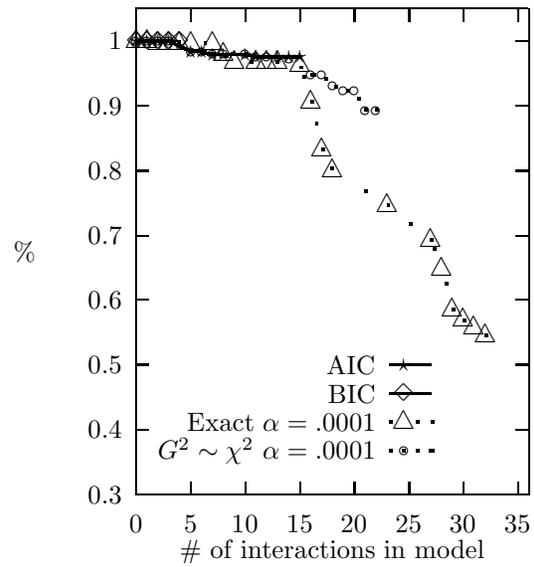


\begin{thebibliography}{}

\bibitem[\protect\citename{Akaike}1974]{Akaike74}
H.~Akaike.
\newblock 1974.
\newblock A new look at the statistical model identification.
\newblock {\em IEEE Transactions on Automatic Control}, AC-19(6):716--723.

\bibitem[\protect\citename{Badsberg}1995]{Badsberg95}
J.~Badsberg.
\newblock 1995.
\newblock {\em An Environment for Graphical Models}.
\newblock {Ph.D.} thesis, Aalborg University.

\bibitem[\protect\citename{Berger \bgroup et al.\egroup }1996]{BergerDD96}
A.~Berger, S.~Della~Pietra, and V.~Della~Pietra.
\newblock 1996.
\newblock A maximum entropy approach to natural language processing.
\newblock {\em Computational Linguistics}, 22(1):39--71.

\bibitem[\protect\citename{Bishop \bgroup et al.\egroup }1975]{BishopFH75}
Y.~Bishop, S.~Fienberg, and P.~Holland.
\newblock 1975.
\newblock {\em Discrete Multivariate Analysis}.
\newblock The MIT Press, Cambridge, MA.

\bibitem[\protect\citename{Black}1988]{Black88}
E.~Black.
\newblock 1988.
\newblock An experiment in computational discrimination of {E}nglish word
  senses.
\newblock {\em IBM Journal of Research and Development}, 32(2):185--194.

\bibitem[\protect\citename{Brown \bgroup et al.\egroup }1991]{BrownPPM91}
P.~Brown, S.~Della~Pietra, and R.~Mercer.
\newblock 1991.
\newblock Word sense disambiguation using statistical methods.
\newblock In {\em Proceedings of the 29th Annual Meeting of the Association for
  Computational Linguistics}, pages 264--304.

\bibitem[\protect\citename{Bruce and Wiebe}1994a]{BruceW94A}
R.~Bruce and J.~Wiebe.
\newblock 1994a.
\newblock A new approach to word sense disambiguation.
\newblock In {\em Proceedings of the ARPA Workshop on Human Language
  Technology}, pages 244--249.

\bibitem[\protect\citename{Bruce and Wiebe}1994b]{BruceW94B}
R.~Bruce and J.~Wiebe.
\newblock 1994b.
\newblock Word-sense disambiguation using decomposable models.
\newblock In {\em Proceedings of the 32nd Annual Meeting of the Association for
  Computational Linguistics}, pages 139--146.

\bibitem[\protect\citename{Bruce \bgroup et al.\egroup }1996]{BruceWP96}
R.~Bruce, J.~Wiebe, and T.~Pedersen.
\newblock 1996.
\newblock The measure of a model.
\newblock In {\em Proceedings of the Conference on Empirical Methods in Natural
  Language Processing}, pages 101--112.

\bibitem[\protect\citename{Dagan \bgroup et al.\egroup }1991]{DaganIS91}
I.~Dagan, A.~Itai, and U.~Schwall.
\newblock 1991.
\newblock Two languages are more informative than one.
\newblock In {\em Proceedings of the 29th Annual Meeting of the Association for
  Computational Linguistics}, pages 130--137.

\bibitem[\protect\citename{Darroch \bgroup et al.\egroup }1980]{DarrochLS80}
J.~Darroch, S.~Lauritzen, and T.~Speed.
\newblock 1980.
\newblock Markov fields and log-linear interaction models for contingency
  tables.
\newblock {\em The Annals of Statistics}, 8(3):522--539.

\bibitem[\protect\citename{Gale \bgroup et al.\egroup }1992]{GaleCY92}
W.~Gale, K.~Church, and D.~Yarowsky.
\newblock 1992.
\newblock A method for disambiguating word senses in a large corpus.
\newblock {\em Computers and the Humanities}, 26:415--439.

\bibitem[\protect\citename{Kreiner}1987]{Kreiner87}
S.~Kreiner.
\newblock 1987.
\newblock Analysis of multidimensional contingency tables by exact conditional
  tests: Techniques and strategies.
\newblock {\em Scandinavian Journal of Statistics}, 14:97--112.

\bibitem[\protect\citename{Leacock \bgroup et al.\egroup }1993]{LeacockTV93}
C.~Leacock, G.~Towell, and E.~Voorhees.
\newblock 1993.
\newblock Corpus-based statistical sense resolution.
\newblock In {\em Proceedings of the ARPA Workshop on Human Language
  Technology}.

\bibitem[\protect\citename{Mooney}1996]{Mooney96}
R.~Mooney.
\newblock 1996.
\newblock Comparative experiments on disambiguating word senses: An
  illustration of the role of bias in machine learning.
\newblock In {\em Proceedings of the Conference on Empirical Methods in Natural
  Language Processing}.

\bibitem[\protect\citename{Ng and Lee}1996]{NgL96}
H.T. Ng and H.B. Lee.
\newblock 1996.
\newblock Integrating multiple knowledge sources to disambiguate word sense: An
  exemplar-based approach.
\newblock In {\em Proceedings of the 34th Annual Meeting of the Society for
  Computational Linguistics}, pages 40--47.

\bibitem[\protect\citename{Pedersen \bgroup et al.\egroup }1996]{PedersenKB96}
T.~Pedersen, M.~Kayaalp, and R.~Bruce.
\newblock 1996.
\newblock Significant lexical relationships.
\newblock In {\em Proceedings of the 13th National Conference on Artificial
  Intelligence}, pages 455--460.

\bibitem[\protect\citename{Schwarz}1978]{Schwarz78}
G.~Schwarz.
\newblock 1978.
\newblock Estimating the dimension of a model.
\newblock {\em The Annals of Statistics}, 6(2):461--464.

\bibitem[\protect\citename{Whittaker}1990]{Whittaker90}
J.~Whittaker.
\newblock 1990.
\newblock {\em Graphical Models in Applied Multivariate Statistics}.
\newblock John Wiley, New York.

\bibitem[\protect\citename{Yarowsky}1993]{Yarowsky93}
D.~Yarowsky.
\newblock 1993.
\newblock One sense per collocation.
\newblock In {\em Proceedings of the ARPA Workshop on Human Language
  Technology}, pages 266--271.

\bibitem[\protect\citename{Zipf}1935]{Zipf35}
G.~Zipf.
\newblock 1935.
\newblock {\em The Psycho-Biology of Language}.
\newblock Houghton Mifflin, Boston, MA.

\end{thebibliography}
\end{document}